\PassOptionsToPackage{table}{xcolor}
\documentclass[sigconf]{acmart}

\setcopyright{none}
\renewcommand\footnotetextcopyrightpermission[1]{}

\AtBeginDocument{%
  }

\usepackage{booktabs}
\usepackage[most]{tcolorbox}
\usepackage[table]{xcolor}

\definecolor{bestcell}{RGB}{224,245,224}
\definecolor{secondcell}{RGB}{255,248,220}
\definecolor{sembidcell}{RGB}{229,238,255}
\definecolor{AIYellowBack}{RGB}{255, 249, 227}
\definecolor{AIYellowFrame}{RGB}{229, 191,  89}
\definecolor{AIGreenTitle}{RGB}{ 22, 125,  90}
\definecolor{AIBoxTitleBlue}{RGB}{31, 62, 122}
\definecolor{AIGrayLine}{RGB}{210, 210, 210}
\newcommand{\PromptSection}[1]{\noindent{\textbf{#1}}\vspace{2pt}}
\newtcolorbox{AIBox}[1][]{%
  enhanced,
  colback=AIYellowBack,
  colframe=AIYellowFrame,
  boxrule=0.6pt,
  arc=2.2mm,
  left=1.8mm,right=1.8mm,top=1.6mm,bottom=1.6mm,
  colbacktitle=AIBoxTitleBlue,
  coltitle=white,
  fonttitle=\bfseries\small,
  attach boxed title to top left={xshift=1.2mm,yshift*=-1mm},
  boxed title style={arc=2.0mm, boxrule=0pt, left=2.2mm,right=2.2mm,top=1.0mm,bottom=1.0mm},
  #1
}

\usepackage{graphicx}
\usepackage{multirow}
\usepackage{enumitem}
\usepackage{algorithm}
\usepackage{float}
\usepackage{algorithmic}
\AtBeginDocument{%
  \hypersetup{
    colorlinks=true,
    linkcolor=blue,
    citecolor=blue,
    urlcolor=blue
  }%
}

\begin{document}

\title{On the Role of Language Representations in Auto-Bidding: Findings and Implications}

\author{%
\parbox{\textwidth}{%
\centering
\textbf{Guanyu Zhu}\textsuperscript{1,2},
\textbf{Jining Luan}\textsuperscript{3},
\textbf{Hanwen Du}\textsuperscript{4},
\textbf{Xinyu Fang}\textsuperscript{5},
\textbf{Sibo Xu}\textsuperscript{6},
\textbf{Ersheng Ni}\textsuperscript{7},
\textbf{Hongji Li}\textsuperscript{8},
\textbf{Jincheng Fang}\textsuperscript{9},
\textbf{Ronghao Chen}\textsuperscript{10},
\textbf{Huacan Wang}\textsuperscript{11},
\textbf{Xuanqi Lan}\textsuperscript{12},
\textbf{Yongxin Ni}\textsuperscript{13},
\textbf{Yiqi Sun}\textsuperscript{14},
\textbf{Youhua Li}\textsuperscript{\textdagger,1}\\[0.8em]
{\normalsize
\textsuperscript{1}City University of Hong Kong \quad
\textsuperscript{2}South China Agricultural University \quad
\textsuperscript{3}University of Electronic Science and Technology of China \quad
\textsuperscript{4}The Ohio State University, Columbus \quad
\textsuperscript{5}Hefei University of Technology \quad
\textsuperscript{6}School of Economics and Management, Wuhan University \quad
\textsuperscript{7}Faculty of Engineering, The University of Queensland \quad
\textsuperscript{8}Mohamed bin Zayed University of Artificial Intelligence \quad
\textsuperscript{9}The Hong Kong University of Science and Technology \quad
\textsuperscript{10}Peking University \quad
\textsuperscript{11}University of Chinese Academy of Sciences \quad
\textsuperscript{12}Santa Clara University \quad
\textsuperscript{13}Boston University \quad
\textsuperscript{14}The University of Hong Kong
}\\[1em]
{\small
\textsuperscript{\textdagger}\textit{Corresponding author.}
}
}%
}

\begin{abstract}
Auto-bidding is a crucial task in real-time advertising markets, where policies must optimize long-horizon value under delivery constraints (e.g., budget and CPA). Existing methods for auto-bidding rely on compact numerical state representations: while they can implicitly capture delivery dynamics, they offer limited support for explicitly representing and controlling high-level intent, evolving feedback, and operator-style strategic guidance in real campaigns. Meanwhile, Large Language Models (LLMs) offer a powerful method for encoding semantic information, it remains unclear when LLMs help and how to integrate them without sacrificing numerical precision. Through systematic preliminary studies, we find that (1) LLM embeddings contain bidding-relevant cues yet cannot replace numerical features, and (2) gains emerge only with careful semantic--numeric integration rather than naive concatenation. Motivated by these findings, we propose \textit{SemBid}, a novel auto-bidding framework that injects LLM-encoded semantics into offline bidding trajectories at the token level. SemBid introduces three semantic inputs: \textit{Task}, \textit{History}, and \textit{Strategy}. It injects these semantics as tokens alongside numerical trajectory tokens and uses self-attention to integrate them, improving controllability and generalization across objectives.
Across diverse scenarios and budget regimes, SemBid outperforms competitive baselines from offline RL and generative sequence modeling, with more consistent gains in overall performance, constraint satisfaction, and robustness.
Our code is available at:
\href{https://github.com/AlanYu04/SemBid-KDD2026}{\textcolor{blue}{here}}.

\end{abstract}

\maketitle
\pagestyle{plain}
\fancyhf{}
\fancyfoot[C]{\thepage}
\renewcommand{\headrulewidth}{0pt}
\fancyfoot[C]{\thepage}
\renewcommand{\headrulewidth}{0pt}

\section{Introduction}

The rapid growth of online advertising has made it a major revenue source for Internet platforms~\cite{gao2024hierrec,wang2015real}. In Real-Time Bidding (RTB), advertisers participate in per-impression auctions, where each impression corresponds to a single opportunity to display an ad to a user~\cite{yuan2014survey,aggarwal2024auto}.
In this process, an advertiser must continuously adjust its bidding strategy under a budget constraint and cost-efficiency requirements~\cite{xu2015smart} to secure as much conversion value as possible in a dynamic traffic and competitive environment~\cite{wu2018budget}. Given the enormous scale of impressions and the temporal fluctuations of market conditions, frequent manual intervention is not only difficult to scale but also often fails to achieve an optimal trade-off between cost and performance. Moreover, advertisers exhibit substantial heterogeneity in business objectives and cost preferences: brand advertising typically emphasizes long-term growth and awareness, pursuing reach under constraints such as per-impression cost; performance advertising, in contrast, prioritizes conversion efficiency and commonly targets constraints such as cost per acquisition (CPA)~\cite{levine2020offline,reid2022can}. Against this backdrop, auto-bidding has emerged as a critical platform capability, requiring continual optimization in non-stationary delivery environments while aligning with diverse preferences to deliver stable and controllable business outcomes.

Deep Reinforcement learning (DRL) has become a mainstream approach for auto-bidding~\cite{cai2017real,aggarwal2024auto,janner2021offline}, spanning early value-learning-based bid optimization methods~\cite{wu2018budget,zhang2016feedback} to more recent offline policy learning paradigms that better match industrial data regimes~\cite{kostrikov2021offline,li2024trajectory,mou2022sustainable}. 
In parallel, generative decision modeling has introduced a new perspective on this problem~\cite{janner2021offline,guo2024aigbgenerativeautobiddingconditional}: Decision Transformer (DT) reformulates policy learning as conditional sequence modeling~\cite{chen2021decision}, predicting actions by conditioning on the desired return, thereby bypassing part of the difficulties in value-function estimation ~\cite{foster2021offline,brandfonbrener2022does} and enabling more effective use of long-range historical context. 
Building on this foundation, DT extensions tailored to bidding scenarios (e.g., GAVE \cite{gao2025generative} and GAS~\cite{li2025gas}) further incorporate mechanisms such as goal modeling, exploration, and search to accommodate complex constraints and evolving preferences~\cite{jiang2025optimal,ding2026c2}.
Nevertheless, existing methods remain largely centered on numerical features: objectives and constraints are typically represented as scalars or low-dimensional vectors, and the resulting policies rely primarily on numerical correlations to implicitly absorb semantics and strategy. There is still substantial room for explicitly representing and controllably leveraging high-level context~\cite{broder2007semantic,zhao2025llm}.

In recent years, advances in large language models (LLMs) for semantic understanding and text generation~\cite{brown2020language} suggest that they can serve as high-level contextual representations ~\cite{reimers2019sentence}for autobid systems~\cite{aggarwal2024auto,edelman2007internet}, complementing information that is difficult to encode with traditional numerical features alone~\cite{nguyen2021multimodal}. In real-world delivery, budget constraint, CPA targets, and stage-wise demands for scaling up or throttling spend are often described and reviewed in natural language. Similarly, operational teams routinely summarize recent performance using language (e.g., ``ROI is low'' or ``cost fluctuates substantially'') and issue strategy directives (e.g., ``raise bids moderately to win more volume'' or ``reduce exploration to control cost''). Such information is not at odds with numerical logs; rather, it captures the decision context at the \textit{objective–feedback–strategy} level~\cite{liu2024dellma}, yet has long remained underutilized by existing models. This motivates a research question: \textit{can language provide actionable contextual signals that make auto-bidding more controllable and stable, while preserving the numerical fidelity required for precise bid optimization?} In Section~\ref{sec:preliminary}, we conduct controlled preliminary experiments to delineate when such language signals are beneficial and to probe effective integration strategies.

Through systematic preliminary experiments, we arrive at three key takeaways. First, LLM embeddings encode rich semantics that are beneficial
for the auto-bidding task. Second, LLM embeddings provide complementary semantics that, when effectively
fused with numerical features, can significantly enhance auto-
bidding performance. Third, while simple fusion methods to combine LLM embeddings and numerical features can be effective, advanced
fusion methods can achieve superior performance by dynamically capturing complex dependencies. These observations directly inform our design choices: Decision Transformer’s token-based formulation provides a natural interface for introducing semantic inputs, and its self-attention can learn to integrate heterogeneous signals in context. Building on this, we propose \textit{SemBid}, which augments Decision Transformer’s original per-step triplet with Task, History, and Strategy, resulting in a sextuple aligned with objective, temporal, and strategic dimensions of decision-making.Concretely, SemBid encodes each semantic signal with a frozen LLM and injects the resulting semantic tokens alongside numerical trajectory tokens into a shared Transformer, which autoregressively predicts the next bid (action) conditioned on both semantics and logged dynamics.
Our contributions are summarized as follows:

\begin{itemize}[leftmargin=*,nosep,topsep=0pt,partopsep=0pt]
    \item Through systematic preliminary experiments, we identify when and how LLM-derived semantics help auto-bidding: semantic cues are informative but cannot replace numerical features, and improvements arise with an advanced fusion mechanism to combine LLM embedding and numerical features.
    \item Motivated by our findings, we design SemBid, a Decision Transformer that introduces Task, History, and Strategy as semantic tokens, encoding campaign intent, recent feedback, and operator priors in a unified sequence model.
    \item We demonstrate consistent improvements across diverse bidding scenarios and evaluation settings (e.g., budget regimes and constraint targets), outperforming a broad range of baselines spanning multiple learning paradigms with stronger overall performance and constraint satisfaction.
\end{itemize}

\section{Related Works}
\subsection{Auto-bidding in Online Advertising}
Auto-bidding strategies have evolved from classical control and online RL methods~\cite{zhang2016feedback,wu2018budget} to offline RL paradigms that learn from logged data, avoiding the cost and risk of live exploration~\cite{cai2017real,jin2018real,fujimoto2019off,kostrikov2021offline}. However, many RL-based formulations rely on the Markov property with compact state abstractions; in auction environments, this design can under-represent longer-term context and temporal dependencies~\cite{allen2021learning,gao2025generative}. Moreover, policies are typically conditioned on numerical summaries (e.g., RTG and a small set of delivery statistics), leaving limited capacity to represent the high-level intent and operator-style guidance that often shape real campaign decisions. To better leverage long histories, recent work has reframed offline policy learning as conditional sequence modeling over trajectories~\cite{janner2021offline}. Decision Transformer (DT) is a seminal representative of this line: it conditions on a target return-to-go (RTG) and autoregressively generates actions along the trajectory~\cite{chen2021decision}. This paradigm has since been adapted to auto-bidding in several ways: GAS introduces a post-training, MCTS-inspired search procedure that aligns a single base policy with heterogeneous advertiser preferences at inference time without retraining~\cite{li2025gas}; GAVE proposes a flexible, score-based RTG formulation coupled with value-guided exploration to handle complex objectives and mitigate OOD issues~\cite{gao2025generative}; and other generative approaches further explore diffusion models for bidding~\cite{guo2024aigbgenerativeautobiddingconditional}. Despite this progress, existing methods still operate primarily on numerical signals, leaving a semantic gap between high-level strategic intent and numerical execution.

\subsection{LLM-Empowered Auto-bidding}

In parallel, RTBAgent explores LLM-based agents for real-time bidding, but emphasizes agentic workflows and tool use rather than integrating semantics into the policy representation itself~\cite{cai2025rtbagent}. Overall, current approaches are either driven by compact numerical conditions (e.g., RTG) with limited access to structured semantic context, or adopt text-centric planning that is not explicitly coupled with the numerical control loop. SemBid bridges these two extremes by retaining a numerically grounded decision model while injecting structured semantic tokens---encoding objectives, feedback, and strategy---directly into the generative decision framework.

\section{Preliminary Study}
\label{sec:preliminary}

Before designing a systematic framework, we run preliminary studies to answer three questions: (1) whether LLM embeddings contain bidding-relevant semantics and to what extent they can substitute numerical features; (2) how different prompt types provide complementary cues; and (3) which fusion mechanisms best integrate semantic and numerical signals. To keep the presentation concise, we report only the main findings here and defer the preliminary-study details to Appendix~\ref{app:preliminary}.

\subsection{Semantic Content in LLM Embeddings}

We first verify whether LLM embeddings encode bidding-relevant information or are merely random noise. We adopt linear probing: if a simple linear model can predict bidding actions from embeddings, the representations contain task-relevant information.

We use Qwen-0.5B (embeddings from other LLMs are also effective for auto-bidding, as will be shown later in Table~\ref{tab:encoder-comparison}) to encode bidding states into natural language descriptions (e.g., ``Budget: 80\%. pValue: High. Time remaining: 2 hours.'') and compare against random baselines. Crucially, we include a shuffled embedding test where sample-embedding correspondences are randomized to eliminate semantic content while preserving statistical properties.

\begin{figure}[t]
\centering
\includegraphics[width=\columnwidth]{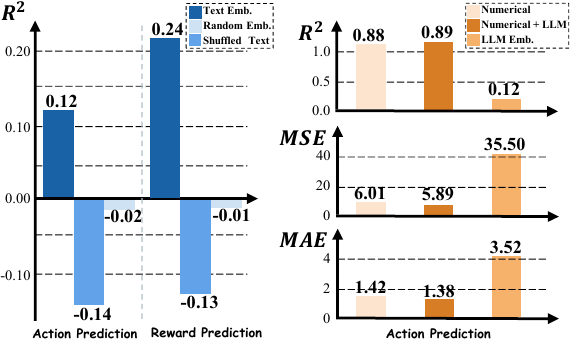}
\vspace{-6mm}
\caption{\textbf{Preliminary results.} \textbf{Left:} Text embeddings predict action/reward better than random or shuffled controls. \textbf{Right:} Semantics alone lag numerical features, but fusion yields a small gain, suggesting careful integration.}
\vspace{-6mm}
\label{fig:preliminary-results}
\end{figure}

State Text Embedding achieves $R^2=0.121$ for action prediction and $R^2=0.241$ for reward prediction (Figure~\ref{fig:preliminary-results}, left), dramatically outperforming random baselines (negative $R^2$ values). Notably, shuffling sample-embedding correspondences drops $R^2$ to near zero (action: -0.015, reward: -0.013), confirming that the semantic content---not statistical patterns---drives prediction. This establishes that LLM embeddings do encode bidding-relevant semantics.

\begin{tcolorbox}[
  enhanced,
  title={Findings 1},
  colframe=blue!40!black,
  colback=blue!2!white,
  fonttitle=\bfseries,
  attach boxed title to top text left={xshift=30mm,yshift=-2.5mm},
  boxed title style={
    size=small,
    colframe=blue!40!black,
    colback=blue!40!black
  }
]
LLM embeddings encode rich semantics that are beneficial for the auto-bidding task.
\end{tcolorbox}

\subsection{Limitations of Semantic-Only Representations}

A natural hypothesis arises from the semantic richness of LLM embeddings: could traditional numerical features be no longer necessary for auto-bidding? Is it possible to rely entirely on semantic representations derived from LLM embeddings?

Results point to two takeaways (Figure~\ref{fig:preliminary-results}, right). First, semantic embeddings alone are insufficient to fully replace numerical features: using only LLM embeddings yields $R^2=0.121$, dramatically below the numerical baseline of $R^2=0.880$ (an 86.25\% relative gap). This is unsurprising in auto-bidding, where decisions hinge on fine-grained arithmetic, including tight budget states, conversion likelihoods, and cost computations. Such signals are difficult to capture with sufficient precision from natural-language descriptions. Second, semantic signals can significantly complement numerical features when integrated appropriately: combining them reaches $R^2=0.886$, a modest but consistent improvement over numerical features alone. Importantly, this gain depends on the fusion mechanism, which we examine in subsequent experiments.

\begin{tcolorbox}[
  enhanced,
  title={Findings 2},
  colframe=blue!40!black,
  colback=blue!2!white,
  fonttitle=\bfseries,
  attach boxed title to top text left={xshift=30mm,yshift=-2.5mm},
  boxed title style={
    size=small,
    colframe=blue!40!black,
    colback=blue!40!black
  }
]
While insufficient on their own, LLM embeddings provide complementary semantics that, when effectively fused with numerical features, can significantly enhance auto-bidding performance.
\end{tcolorbox}

\subsection{Fusion Mechanisms for Semantic Signals}

Finally, we investigate how to effectively integrate semantic embeddings with numerical features. We compare simple concatenation against more sophisticated fusion mechanisms.

\begin{table}[t]
\caption{Fusion Mechanism Comparison}
\label{tab:preliminary-fusion}
\centering
{\small
\setlength{\tabcolsep}{6pt}
\renewcommand{\arraystretch}{1.05}
\begin{tabular}{lcc}
\toprule
Fusion Method & $R^2$ & \textit{v.s.} Baseline \\
\midrule
Numerical Only (Baseline) & 0.880 & - \\
\midrule
Concat Fusion & 0.850 & -3.41\% \\
Residual Fusion & 0.872 & -0.91\% \\
Gated Fusion & 0.863 & -1.93\% \\
\textbf{Cross-Attention Fusion} & \textbf{0.886} & \textbf{+0.68\%} \\
FiLM Fusion & 0.859 & -2.39\% \\
\bottomrule
\end{tabular}
}
\end{table}

Simple concatenation performs worst, reducing $R^2$ by 3.41\%, which suggests that treating embeddings as ordinary features can dilute strong numerical signals. In contrast, attention-based fusion performs best: cross-attention achieves $R^2=0.886$, slightly surpassing the numerical baseline. This indicates that selective fusion mechanisms are better at filtering when semantic context is useful and when it should be ignored. These results motivate our attention-based integration in SemBid, where the transformer can learn context-dependent weighting between semantic tokens and numerical trajectory tokens.
\begin{tcolorbox}[
  enhanced,
  title={Findings 3},
  colframe=blue!40!black,
  colback=blue!2!white,
  fonttitle=\bfseries,
  attach boxed title to top text left={xshift=30mm,yshift=-2.5mm},
  boxed title style={
    size=small,
    colframe=blue!40!black,
    colback=blue!40!black
  }
]
While simple fusion methods can be effective, advanced fusion methods can achieve superior performance by dynamically capturing complex dependencies.
\end{tcolorbox}

\section{Method}

\begin{figure*}[t]
\centering
\includegraphics[width=\textwidth]{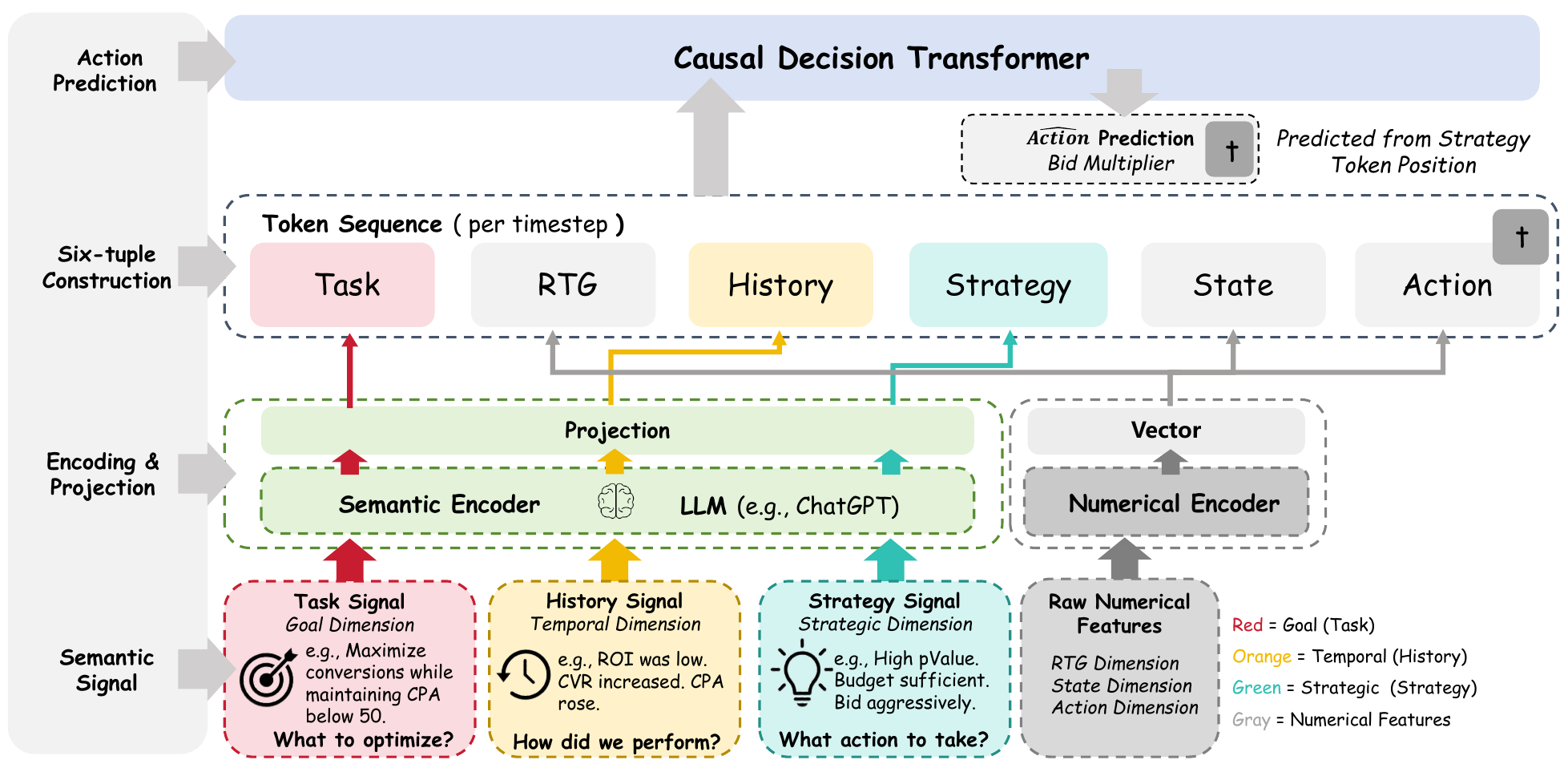}
\caption{Overview of the SemBid framework. The model augments Decision Transformer with three complementary semantic signals, i.e., Task, History, and Strategy  dimensions. Semantic signals are encoded by a frozen LLM and fed as token embeddings to the Causal Transformer. Action is predicted from the State token position.}
\label{fig: framework}
\end{figure*}

The overall framework of SemBid is illustrated in Figure~\ref{fig: framework}. 
We augment numerical bidding trajectories with three semantic signals---\textsc{Task}, \textsc{History}, and \textsc{Strategy}---encoded by a frozen LLM. 
Compared to prior DT-style bidding models that condition mainly on numerical tokens (e.g., RTG and delivery statistics), SemBid injects explicit semantic priors to contextualize the state, and learns an autoregressive policy under a causal mask.

\subsection{Problem Statement}

\subsubsection{Auto-bidding as Constrained Sequential Optimization}

We consider an advertiser participating in Real-Time Bidding (RTB) auctions over a finite campaign horizon. Consistent with standard industrial practices, we adopt a period-level control framework to balance decision granularity with computational efficiency. The horizon is discretized into $T$ consecutive time steps. At each step $t \in \{1, \dots, T\}$, the system manages a batch of $N_t$ impression opportunities, indexed by the set $\mathcal{I}_t$.

\paragraph{Bidding Mechanism.}
For each impression (i.e., a single opportunity to display an ad to a user) $n \in \mathcal{I}_t$, let $v_{t,n} \in \mathbb{R}_{+}$ denote the predicted conversion value. The agent controls a policy parameterized by a scalar multiplier $\lambda_t \in [0, \lambda_{\max}]$, which determines the bid $b_{t,n} = \lambda_t \cdot v_{t,n}$. We assume a Generalized Second-Price (GSP) auction environment. The advertiser wins impression $n$ ($x_{t,n}=1$) if $b_{t,n}$ exceeds the highest competing bid, paying the market price $c_{t,n}$. Let $y_{t,n} \in \{0,1\}$ denote the ground-truth conversion outcome. The period-level total value ($r_t$), conversion count ($cv_t$), and spend ($sp_t$) are aggregated as:
\begin{equation}
\begin{aligned}
    r_t = \sum_{n \in \mathcal{I}_t} x_{t,n} v_{t,n} \quad
    cv_t = \sum_{n \in \mathcal{I}_t} x_{t,n} y_{t,n} \quad
    sp_t = \sum_{n \in \mathcal{I}_t} x_{t,n} c_{t,n}
\end{aligned}
\label{eq:period-agg}
\end{equation}

\paragraph{Constrained Objective.}
The advertiser aims to maximize the total conversion value over the campaign horizon while satisfying operational constraints. Formally, we solve
\begin{equation}
\begin{aligned}
\max_{\lambda_1, \lambda_2,\cdots,\lambda_T}  \sum_{t=1}^{T} r_t \quad \text{s.t.} \quad g_j(\tau) \leq C_j\quad j = 1,\ldots,J
\end{aligned}
\end{equation}
\noindent where $g_j(\cdot)$ specifies constraint $j$ as a function of the full trajectory $\tau$.
In this work, we focus on two standard constraints in RTB:
(i) the Budget Constraint $g_1(\tau) = \sum_{t=1}^{T} sp_t \leq B$
and (ii) the cost-per-acquisition (CPA) constraint
\begin{equation}
g_2(\tau)=\frac{\sum_{t=1}^{T} sp_t}{\sum_{t=1}^{T} cv_t + \epsilon} \qquad g_2(\tau)\leq C_{\text{CPA}}
\end{equation}
\noindent where $\epsilon>0$ is a small constant to avoid division by zero. Both constraints are evaluated at the trajectory level over the full horizon. 

\subsubsection{Offline Reinforcement Learning Formulation}

We cast the constrained bidding problem as a finite-horizon Markov Decision Process (MDP) defined by the tuple $(\mathcal{S}, \mathcal{A}, \mathcal{P}, \mathcal{R})$, where $\mathcal{S}$ denotes the state space, $\mathcal{A}$ denotes the action space, $\mathcal{P}:\mathcal{S}\times\mathcal{A}\rightarrow\mathcal{S}$ denotes the transition to the next state after taking an action from the current state, and $\mathcal{R}:\mathcal{S}\times\mathcal{A}\rightarrow\mathbb{R}$ denotes the reward function. At each timestep $t \in \{1,\ldots,T\}$, the agent observes a state $s_t \in \mathcal{S} \subseteq \mathbb{R}^{d_s}$ summarizing campaign status (e.g., remaining budget, time elapsed, and historical performance metrics), and selects an action $a_t = \lambda_t \in [0, \lambda_{\max}]$ corresponding to the bid multiplier. The environment then transitions to $s_{t+1}$ and returns a reward $r_t$, where $r_t$ is the period-level total conversion value defined in Eq.~\ref{eq:period-agg}. Following~\cite{chen2021decision}, we compute the return-to-go $G_t=\sum_{\tau=t}^{T} r_\tau$ and use $G_t$ as the return-conditioning token during training.

We strictly follow the offline RL setting, where we learn policies solely from a static dataset $\mathcal{D}$ without online interaction. To align with the sequence modeling paradigm of Decision Transformer, we formulate the dataset as a collection of \textbf{trajectories}:
\begin{equation}
    \mathcal{D} = \{ \tau_i \}_{i=1}^{N}, \quad \text{where } \tau_i = (s_1, a_1, r_1, \dots, s_T, a_T, r_T)
\end{equation}
This trajectory-based formulation treats auto-bidding as a sequence generation problem, where the goal is to predict optimal actions conditioned on historical context.

\subsection{Semantic Signal Design}
\label{sec:semantic_signals}
\begin{algorithm}[t]
\caption{Training procedure of SemBid}
\label{alg:sembid}
\begin{algorithmic}[1]
\REQUIRE Offline trajectories $\mathcal{D}$; prompt templates $\{f_k\}_{k\in\mathcal{K}}$ with $\mathcal{K}=\{\textsc{task},\textsc{hist},\textsc{strat}\}$;
frozen LLM encoder $\mathcal{E}$; semantic projections $\{W_k\}_{k\in\mathcal{K}}$;
numerical encoders $\{\phi_{\textsc{rtg}},\phi_s,\phi_a\}$; Decision Transformer $DT_{\theta}$; action head $W_{\text{out}}$.
\ENSURE Trained parameters $\theta$.

\STATE Initialize $\theta$
\FOR{training step $=1$ to MaxStep}
    \STATE Sample a mini-batch of trajectories $\{\tau\}$ from $\mathcal{D}$
    \FOR{each trajectory $\tau$ in the mini-batch}
        \STATE Compute return-to-go $\{G_t\}_{t=1}^{T}$ from rewards $\{r_t\}_{t=1}^{T}$
        
        \STATE \textbf{(Batched token construction for all $t$)}:
        $\ell_t^{k}\!\leftarrow\! f_k(c_t)$ and
        $\mathbf{v}_t^{k}\!\leftarrow\! W_k\,\mathcal{E}(\ell_t^{k}),\; \forall k\!\in\!\mathcal{K},\forall t$
        \STATE $\mathbf{v}_t^{\textsc{rtg}}\!\leftarrow\!\phi_{\textsc{rtg}}(G_t),\;
               \mathbf{v}_t^{s}\!\leftarrow\!\phi_s(s_t),\;
               \mathbf{v}_t^{a}\!\leftarrow\!\phi_a(a_t),\; \forall t$
        \STATE $\mathbf{x}_t \leftarrow
        \big[\mathbf{v}_t^{\textsc{task}},\,\mathbf{v}_t^{\textsc{rtg}},\,\mathbf{v}_t^{\textsc{hist}},\,\mathbf{v}_t^{\textsc{strat}},\,\mathbf{v}_t^{s},\,\mathbf{v}_t^{a}\big],\; \forall t$
        \STATE $\mathbf{X}\leftarrow(\mathbf{x}_1,\ldots,\mathbf{x}_T)$ with a causal mask
        
        \STATE Predict actions at state-token positions:
        $\hat{a}_t \leftarrow W_{\text{out}}\,DT_{\theta}(\mathbf{X})^{(t,\textsc{state})}$
        \STATE $\mathcal{L} \leftarrow \sum_{t} m_t \cdot \|\hat{a}_t-a_t\|_2^2$
        \STATE Update $\theta$ by minimizing $\mathcal{L}$
    \ENDFOR
\ENDFOR
\STATE \textbf{return} $\theta$
\end{algorithmic}
\end{algorithm}

To augment the numerical state representation with high-level context, we introduce three complementary semantic signals corresponding to the \textbf{goal}, \textbf{temporal}, and \textbf{strategic} dimensions. Each signal is generated from the bidding context via template-based verbalization and encoded by a frozen LLM encoder $\mathcal{E}(\cdot)$:

\begin{itemize}[noitemsep,topsep=3pt,leftmargin=*]

    \item \textbf{Task (Goal Dimension)}: Explicitly encodes the campaign objective and hard constraints (e.g., "Maximize conversions while keeping CPA $\leq$ 50").

    \item \textbf{History (Temporal Dimension)}: Summarizes recent performance trends and feedback to contextualize the current state (e.g., "CPA is drifting upward").

    \item \textbf{Strategy (Strategic Dimension)}: Provides high-level heuristic guidance to ground the decision-making (e.g., "Budget is sufficient; bid aggressively").

\end{itemize}

We formally define each signal below.

\paragraph{Task Signal.}The vector $\mathbf{v}^{task}_t$ explicitly encodes the optimization objectives and hard constraints. It serves as a global anchor for the decision process:
\begin{equation}
    \mathbf{v}^{task}_t = \mathcal{E}\!\left(f_{\textsc{task}}(c_t)\right)\quad
\end{equation}
\paragraph{History Signal.} The vector $\mathbf{v}^{hist}_t$ summarizes historical performance metrics. In practice, we employ a template function $f_{\text{hist}}$ to verbalize short-term performance dynamics (e.g., describing trends in CPA or CVR deviations):
\begin{equation}
    \mathbf{v}^{hist}_t = \mathcal{E}(f_{\text{hist}}(\text{ROI}_{<t}, \text{CVR}_{<t}, \text{CPA}_{<t}))
\end{equation}

\paragraph{Strategy Signal (Strategic Dimension).}
The vector $\mathbf{v}^{strat}_t$ provides high-level heuristic guidance to ground the decision-making. We use a function $f_{\text{strat}}$ to generate actionable advice based on the current opportunity quality ($p_t$) and budget status ($B_t$):
\begin{equation}
    \mathbf{v}^{strat}_t = \mathcal{E}(f_{\text{strat}}(p_t, B_t))
\end{equation}

\subsection{SemBid Architecture}

We implement SemBid using a causal Transformer backbone, extending the standard Decision Transformer (DT) architecture to accommodate high-dimensional semantic signals. The architectural novelty lies in the construction of the sextuplet input sequence. By carefully designing the causal ordering of tokens, we enforce a dependency structure where the model's decision-making is explicitly grounded in the preceding strategic context.

\paragraph{Feature Embedding.}
To align heterogeneous inputs into a unified latent space $\mathbb{R}^d$, we employ two parallel encoding pathways before sequence construction.
\begin{itemize}[leftmargin=*]
    \item \textbf{Semantic Encoder:} For each signal type $k \in \{\textsc{task}, \textsc{hist}, \textsc{strat}\}$, the raw text is first encoded by the frozen LLM into a dense vector $\mathbf{z}_t^{k}$. We then apply a learnable linear projection $W_k$ to obtain the token embedding:
    \begin{equation}
        \mathbf{v}_t^{k} = W_k \mathbf{z}_t^{k}
    \end{equation}
    \item \textbf{Numerical Encoder:} Similarly, we embed numerical inputs (return-to-go $G_t$, state $s_t$, and action $a_t$) using dedicated MLP encoders:
    \begin{equation}
        \mathbf{v}_t^{\textsc{rtg}} = \phi_{\textsc{rtg}}(G_t),\quad
        \mathbf{v}_t^{s} = \phi_s(s_t),\quad
        \mathbf{v}_t^{a} = \phi_a(a_t).
    \end{equation}

\end{itemize}

\paragraph{Sextuplet Token Construction.}
Different from the standard triplet $(R, s, a)$ in Decision Transformer, we construct a sextuplet sequence at each timestep $t$. We explicitly position the semantic signals \textit{before} the numerical state to ensure they serve as conditioning priors. The input sequence is defined as:
\begin{equation}
    \mathbf{x}_t =
    \left[
    \mathbf{v}_t^{\textsc{task}},
    \mathbf{v}_t^{\textsc{rtg}},
    \mathbf{v}_t^{\textsc{hist}},
    \mathbf{v}_t^{\textsc{strat}},
    \mathbf{v}_t^{s},
    \mathbf{v}_t^{a}
    \right].
\end{equation}
where $\mathbf{v}_t^{\textsc{task}}, \mathbf{v}_t^{\textsc{hist}}, \mathbf{v}_t^{\textsc{strat}}$ are the semantic embeddings, and $\mathbf{v}_t^{\textsc{rtg}}, \mathbf{v}_t^{s}, \mathbf{v}_{t}^{a}$ are the numerical embeddings. Note that $\mathbf{v}_{t}^{a}$ represents the ground-truth action at timestep $t$. During training, we employ a causal mask to prevent information leakage, ensuring that the prediction of $\hat{a}_t$ depends solely on the tokens preceding $\mathbf{v}_{t}^{a}$ (i.e., ending at the state token $\mathbf{v}_t^{s}$).
\noindent\textbf{Causal policy modeling.}
We model the bidding policy as
\begin{equation}
a_t \sim \pi_\theta(\cdot \mid \mathbf{x}_{\le t,<a})
= DT_\theta(\cdot \mid
\mathbf{v}^{\textsc{task}}_{\le t},
\mathbf{v}^{\textsc{rtg}}_{\le t},
\mathbf{v}^{\textsc{hist}}_{\le t},
\mathbf{v}^{\textsc{strat}}_{\le t},
\mathbf{v}^{s}_{\le t})
\label{eq:policy-causal}
\end{equation}
where $\mathbf{x}_{\le t,<a}$ denotes the token prefix up to (and including) the state token at time $t$.
In implementation, action tokens are included only as shifted targets and are masked out when predicting at the state-token positions.

\paragraph{Semantic-Numerical Fusion.}
The Transformer processes the concatenated trajectory $\tau = (\mathbf{x}_1, \ldots, \mathbf{x}_T)$ under a causal mask. Crucially, because the \textbf{Strategy} token precedes the \textbf{State} token, the self-attention mechanism forces the state representation to aggregate the strategic guidance before decision-making.

We extract the contextualized output at the state-token position, denoted as $\mathbf{h}_t^{s}$, and predict the bid multiplier via a linear head:
\begin{equation}
    \mathbf{h}_t^{s} = \text{Transformer}(\tau)^{(t, \text{state})} \quad \hat{a}_t = W_{\text{out}} \mathbf{h}_t^{s}
\label{eq:action_pred}    
\end{equation}
This design ensures that the predicted action $\hat{a}_t$ is mathematically conditioned on the full semantic context fused into the state representation. The complete training pipeline, including the semantic token construction, causal masking, and the optimization objective, is summarized in Algorithm~\ref{alg:sembid}.

\section{Experiments}

Our experiments are designed to address the following questions:
\begin{itemize}[noitemsep,nolistsep,leftmargin=*]
\item \textbf{RQ1}: Does \textit{SemBid} outperform strong baselines across budget regimes and conversion-sparsity levels?
\item \textbf{RQ2}: How do the task, history, and strategy semantic tokens contribute to SemBid, and how do they interact under ablation?
\item \textbf{RQ3}: How sensitive is SemBid to prompt design and LLM encoder choice?
\end{itemize}

\subsection{Experimental Setup}
\subsubsection*{Environment and datasets.}
We evaluate on AuctionNet~\cite{su2024auctionnet}, a high-fidelity real-time bidding simulator that captures market stochasticity and multi-agent competition under a controlled, fully reproducible setup.
Unlike static logged datasets, AuctionNet allows us to vary key environment factors while keeping the auction mechanism and evaluation pipeline fixed.
Our stress tests follow two orthogonal settings: \textbf{(i)} three benchmark subsets that represent distinct auto-bidding failure modes, and \textbf{(ii)} five budget regimes that vary the delivery pressure.

\noindent \textbf{Subset construction.}
We instantiate three subsets from AuctionNet based on conversion density and constraint tightness:
\begin{itemize}[noitemsep,nolistsep,leftmargin=*]
\item \textbf{AuctionNet-High (constraint-dominant):} Dense conversions with tight CPA constraints; the key challenge is precise control to avoid violations rather than finding opportunities.
\item \textbf{AuctionNet-Medium (pacing-dominant):} Typical market dynamics with moderate conversions; the main difficulty is pacing spend to balance delivery and budget longevity.
\item \textbf{AuctionNet-Low (exploration-dominant):} Extremely sparse rewards with distribution shift; performance hinges on robust exploration when credit assignment is unreliable.
\end{itemize}

\noindent \textbf{Budget settings.}
For each subset, we scale the nominal per-episode budget $B_0$ by $\rho \in \{0.5, 0.75, 1.0, 1.25, 1.5\}$ (i.e., $B=\rho B_0$).
Smaller $\rho$ induces budget-scarce delivery, while larger $\rho$ emphasizes pacing and constraint control.
Unless otherwise stated, all other environment settings are fixed across budget regimes.

\subsubsection*{Evaluation Metrics.}
We follow AuctionNet~\cite{su2024auctionnet} and use \textbf{Score} as the sole evaluation metric.
Score measures the achieved value while penalizing CPA violations via a multiplicative term.
Let $x_i\in\{0,1\}$ denote whether impression $i$ is won, with cost $c_i$ and conversion indicator $y_i\in\{0,1\}$.
We set the value per conversion to $1$, i.e., $v_i = y_i$, and define
\begin{equation}
V = \sum_i x_i v_i,\quad C=\sum_i x_i c_i,\quad \mathrm{CPA}=\frac{C}{V+10^{-10}}.
\end{equation}
The penalty is
\begin{equation}
\text{penalty} = \min\left\{\left(\frac{C_{\textsc{cpa}}}{\mathrm{CPA}}\right)^{\beta},\,1\right\},\quad \beta=2,
\end{equation}
and the final score is
\begin{equation}
\text{Score} = V\cdot \text{penalty}.
\end{equation}

Budget is enforced by the evaluation environment (e.g., dropping bids when the remaining budget is insufficient), rather than being included in the score term.

\subsubsection*{Baselines.}
We compare SemBid with representative baselines covering classical feedback control, offline RL, imitation learning, and DT-style generative bidding:
\begin{itemize}[noitemsep,nolistsep,leftmargin=*]
\item \textbf{PID}~\cite{zhang2016feedback}: a feedback controller that adjusts bids to track budget/CPA targets.
\item \textbf{BC}~\cite{torabi2018behavioral}: behavior cloning that directly imitates logged bidding actions.
\item \textbf{BCQ}~\cite{fujimoto2019off}: offline RL with action-space constraints to stay close to the data support.
\item \textbf{IQL}~\cite{kostrikov2021offline}: expectile-based offline RL that avoids explicit OOD action evaluation.
\item \textbf{CQL}~\cite{kumar2020conservative}: conservative offline RL that penalizes over-optimistic Q-values.
\item \textbf{TD3+BC} ~\cite{fujimoto2021minimalist}: an actor--critic baseline with a behavior-cloning regularizer for offline stability.
\item \textbf{DT}~\cite{chen2021decision}: a transformer policy trained via return-conditioned sequence modeling.
\item \textbf{GAS}~\cite{li2025gas}: a DT-based method augmented with post-training search.
\item \textbf{GAVE}~\cite{gao2025generative}: a value-guided generative bidding method for improved action exploration.
\end{itemize}
\noindent\textbf{Reproducibility details.}
Dataset statistics and environment characterization are in Appendix~\ref{app:dataset}; baseline configurations are in Appendix~\ref{app:baselines}; \textbf{prompt-template variants (different prompt styles/formulations) for semantic signals} are in Appendix~\ref{app:semantic_prompt_templates}; and preliminary-study details are in Appendix~\ref{app:preliminary}.

\begin{table*}[htbp]
\caption{AuctionNet performance under different budget settings. \underline{Underline}: second best; \textbf{bold}: best. \textbf{vs GAVE} and \textbf{vs DT} are relative improvements over GAVE and DT, respectively. *denotes a significant improvement over the best baseline (two-sided t-test, $p<0.05$).}
\label{tab:main-results}
\centering
\setlength{\tabcolsep}{2.5pt}
\resizebox{\textwidth}{!}{
\begin{tabular}{l|c|ccccccccc|ccc}
\toprule
Dataset & Budget & PID & BCQ & IQL & BC & TD3+BC & CQL & DT & GAS & GAVE & \textbf{SemBid} & \textbf{vs.\ GAVE} & \textbf{vs.\ DT} \\
\midrule
\multirow{5}{*}{AuctionNet-High}
& 50\%  & 74.65  & 34.66  & \underline{230.86} & 178.56 & 215.23 & 221.89 & 188.72 & 224.50 & 223.24 & \textbf{238.20}$^{*}$ & +6.7\% & +26.2\% \\
& 75\%  & 86.05  & 59.71  & 301.40 & 277.67 & 306.03 & \underline{314.97} & 282.35 & 302.40 & 301.60 & \textbf{320.60}$^{*}$ & +6.3\% & +13.5\% \\
& 100\% & 133.59 & 118.72 & 350.93 & 355.96 & 371.76 & 377.79 & 356.14 & 390.10 & \underline{392.20} & \textbf{412.20}$^{*}$ & +5.1\% & +15.7\% \\
& 125\% & 164.93 & 167.27 & 394.88 & 426.55 & 457.78 & 425.77 & 430.04 & 443.60 & \underline{459.26} & \textbf{474.87}$^{*}$ & +3.4\% & +10.4\% \\
& 150\% & 198.73 & 182.13 & 437.26 & 490.19 & 503.87 & 492.79 & 469.45 & 498.80 & \underline{518.93} & \textbf{541.76}$^{*}$ & +4.4\% & +15.4\% \\
\midrule
\multirow{5}{*}{AuctionNet-Medium}
& 50\%  & 39.07  & 170.28 & 169.38 & \underline{182.20} & 174.35 & 180.70 & 139.70 & 175.50 & 173.24 & \textbf{182.77}$^{*}$ & +5.5\% & +30.8\% \\
& 75\%  & 65.66  & 224.85 & 208.29 & 218.20 & 229.06 & 231.76 & 209.00 & \underline{240.20} & 240.08 & \textbf{252.80}$^{*}$ & +5.3\% & +21.0\% \\
& 100\% & 98.08  & 265.60 & 229.31 & 236.27 & 246.91 & 258.82 & 263.50 & \underline{298.50} & 296.55 & \textbf{310.19}$^{*}$ & +4.6\% & +17.7\% \\
& 125\% & 132.30 & 295.81 & 239.90 & 270.30 & 263.00 & 282.21 & 318.20 & 334.80 & \underline{337.48} & \textbf{349.63}$^{*}$ & +3.6\% & +9.9\% \\
& 150\% & 170.92 & 310.39 & 246.06 & 283.18 & 267.62 & 325.04 & 347.44 & 358.50 & \underline{369.14} & \textbf{378.37}$^{*}$ & +2.5\% & +8.9\% \\
\midrule
\multirow{5}{*}{AuctionNet-Low}
& 50\%  & 11.11  & 16.21 & \underline{17.96} & 15.96 & \textbf{18.07} & 17.71 & 15.90 & 17.40 & 17.00 & 17.90$^{*}$ & +5.3\% & +12.6\% \\
& 75\%  & 15.98  & 24.52 & \textbf{25.23} & 21.38 & 18.79 & 23.99 & 23.70 & 24.60 & 24.37 & \underline{25.10}$^{*}$ & +3.0\% & +5.9\% \\
& 100\% & 20.88  & 31.00 & 31.09 & 26.75 & 20.84 & 28.61 & 30.00 & \underline{31.20} & 31.10 & \textbf{32.00}$^{*}$ & +2.9\% & +6.7\% \\
& 125\% & 25.91  & 36.20 & 36.24 & 30.72 & 21.05 & 33.28 & 36.10 & 37.30 & \underline{38.22} & \textbf{39.21}$^{*}$ & +2.6\% & +8.6\% \\
& 150\% & 29.89  & 42.02 & 40.85 & 36.37 & 21.30 & 36.15 & 39.40 & 41.80 & \underline{42.39} & \textbf{43.03}$^{*}$ & +1.5\% & +9.2\% \\
\bottomrule
\end{tabular}
}
\end{table*}

\subsection{Overall Performance (RQ1)}

Semantic augmentation yields consistent improvements, with the largest gains emerging in informative-feedback regimes (AuctionNet -High/Medium), where bidding requires long-horizon pacing and tight constraint trade-offs. Table~\ref{tab:main-results} reports overall results on AuctionNet-High/Medium/Low under five budget settings (50\%--150\%).

\noindent\textbf{Robust gains across budgets and settings.}
SemBid achieves the best score in 13 out of 15 configurations. At the standard 100\% budget, SemBid reaches 412.20 (High), 310.19 (Medium), and 32.00 (Low), outperforming both classical offline RL baselines and recent generative methods (DT/GAS/GAVE). The only exceptions occur on the most reward-sparse subset (AuctionNet-Low) under scarce budgets (50\% and 75\%), where TD3+BC and IQL marginally outperform SemBid. A plausible explanation is that in extremely sparse-feedback regimes, more conservative offline baselines can be competitive by staying closer to the logged support, leaving limited headroom for semantic guidance to differentiate actions.

\begin{figure}[t]
    \centering
    \includegraphics[width=0.48\textwidth]{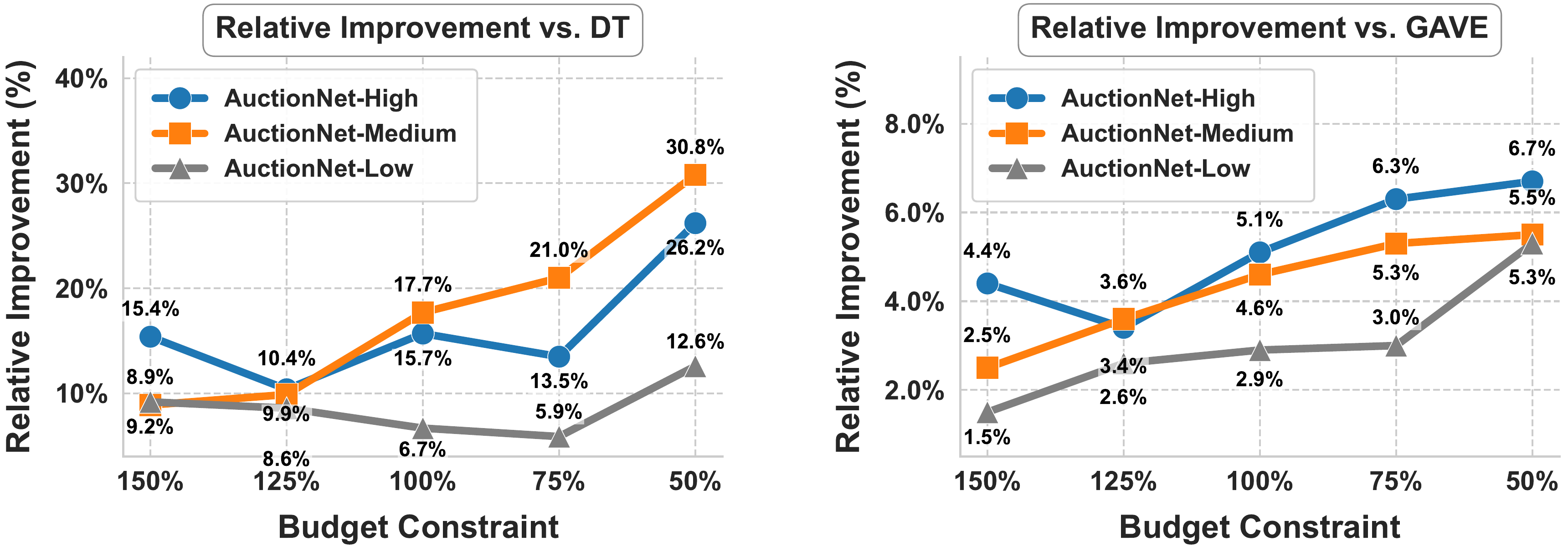}
    \vspace{-4mm}
    \caption{Average relative gains of SemBid under different reward sparsity levels (High$\rightarrow$Low) and different budget levels (150\% $\rightarrow$ 50\%), measured against DT and GAVE.}
    
    \label{fig:sparsity-gain}
    \vspace{-5mm}
\end{figure}

\noindent\textbf{Semantics help most when decision-making requires structured trade-offs.}
We find that \textit{SemBid's advantage is largest in conversion-rich regimes}, as summarized in Figure~\ref{fig:sparsity-gain}. 
At the 100\% budget setting, SemBid improves over vanilla DT by +15.7\% on AuctionNet-High and +17.7\% on AuctionNet-Medium, compared to a smaller gain of +6.7\% on AuctionNet-Low.
This pattern suggests that semantic tokens are most beneficial when the policy must resolve \emph{structured, long-horizon trade-offs}---budget pacing, CPA control, and strategic aggressiveness---under sufficiently informative feedback.
In extremely sparse settings (AuctionNet-Low), the return signal is weak, and trajectories provide limited evidence to validate or refine high-level guidance, so conservative offline baselines remain competitive and semantic augmentation yields smaller incremental gains.
Moreover, as the budget becomes less scarce, the agent has more room to maneuver, and semantic conditioning can more effectively shape pacing and constraint-aware strategies, leading to larger absolute improvements. A similar trend holds against the strongest non-SemBid baseline in each setting. At 100\% budget, SemBid improves over the best competitor by +5.1\% (High), +4.6\% (Medium), and +2.9\% (Low), confirming that the margin over strong baselines also benefits from semantic augmentation across difficulty levels.

\subsection{Ablation Studies (RQ2)}

On AuctionNet-High (100\% budget), strategy provides the largest marginal gain; its effectiveness is amplified by task grounding, while history contributes more modestly. We ablate the three semantic dimensions on AuctionNet-High at 100\% budget. Table~\ref{tab:ablation} reports average score changes when removing each token type and their combinations.

\begin{table}[h]
\caption{Ablation on AuctionNet-High (100\% budget) by removing semantic tokens (task, history, strategy). Avg Score is the mean performance; Rel. to Full is normalized to the full SemBid; Delta is the drop from full.}
\label{tab:ablation}
\vspace{-2mm}
\begin{center}
\begin{small}
\setlength{\tabcolsep}{4pt}
\renewcommand{\arraystretch}{0.95}
\resizebox{\columnwidth}{!}{%
\begin{tabular}{lccc}
\toprule
Ablation Type & Avg Score & Rel. to Full & Delta \\
\midrule
\textit{full} \textbf{SemBid} & \textbf{412.20} & 100\% & - \\
\midrule
\textit{w.o.} task & 387.52 & 94.0\% & -24.68 \\
\textit{w.o.} history & 400.25 & 97.1\% & -11.95 \\
\textit{w.o.} strategy & 365.13 & 88.6\% & -47.07 \\
\midrule
\textit{w.o.} task\_history & 392.93 & 95.3\% & -19.27 \\
\textit{w.o.} task\_strategy & 354.92 & 86.1\% & -57.28 \\
\textit{w.o.} history\_strategy & 366.69 & 89.0\% & -45.51 \\
\midrule
\textit{w.o.} all (Vanilla DT) & 356.14 & 86.4\% & -56.06 \\
\bottomrule
\end{tabular}
}
\end{small}
\end{center}
\vspace{-6mm}
\end{table}
\noindent\textbf{Strategy dominates because it directly conditions action prediction.}
Among the three tokens, \textit{strategy} is the most critical: dropping it reduces the score from 412.20 to 365.13 ($-47.07$), eliminating most of the improvement over vanilla DT. In comparison, removing \textit{task} leads to a moderate drop to 387.52 ($-24.68$), while removing \textit{history} has the smallest effect (400.25, $-11.95$). This ranking is not coincidental---it reflects the causal structure of SemBid's sextuplet sequence. In the causal attention mask, the strategy token is positioned immediately before the state token from which actions are predicted (Eq.~\ref{eq:action_pred}). It therefore serves as the most proximal semantic conditioning signal for action generation, directly encoding actionable guidance (e.g., ``bid aggressively'' vs.\ ``conserve budget'') that shapes the decision boundary. History, by contrast, encodes temporal trends that substantially overlap with the 4-dimensional history features already present in the numerical state vector (Appendix~\ref{app:state-features}), explaining its comparatively smaller marginal contribution.

\noindent\textbf{Task grounds strategy in the right objective context.}
The pairwise ablations reveal a critical interaction: removing both task and strategy (\textit{w.o.\ task\_strategy}) yields 354.92, which is slightly \textit{below} vanilla DT (356.14) and markedly worse than removing strategy alone (365.13). This indicates that without task information, the remaining semantic signals can actually \textit{mislead} the model---strategy cues like ``bid aggressively'' become ambiguous without the objective anchor that task provides (e.g., whether the goal is CPA control or volume maximization). In contrast, removing task and history (\textit{w.o.\ task\_history}) results in 392.93, remaining close to the full model. This asymmetry confirms a clear design principle: \textbf{strategy provides the largest marginal gain, but task is essential for grounding its interpretation}. In conclusion, these two signals contribute most to the performance improvement of SemBid over vanilla DT.

\subsection{Sensitivity to Semantic Signal Design (RQ3)}

\subsubsection{Sensitivity to Prompt Variants}

\begin{figure}[htbp]
\vspace{-3mm}

\centering
\includegraphics[width=0.48\textwidth]{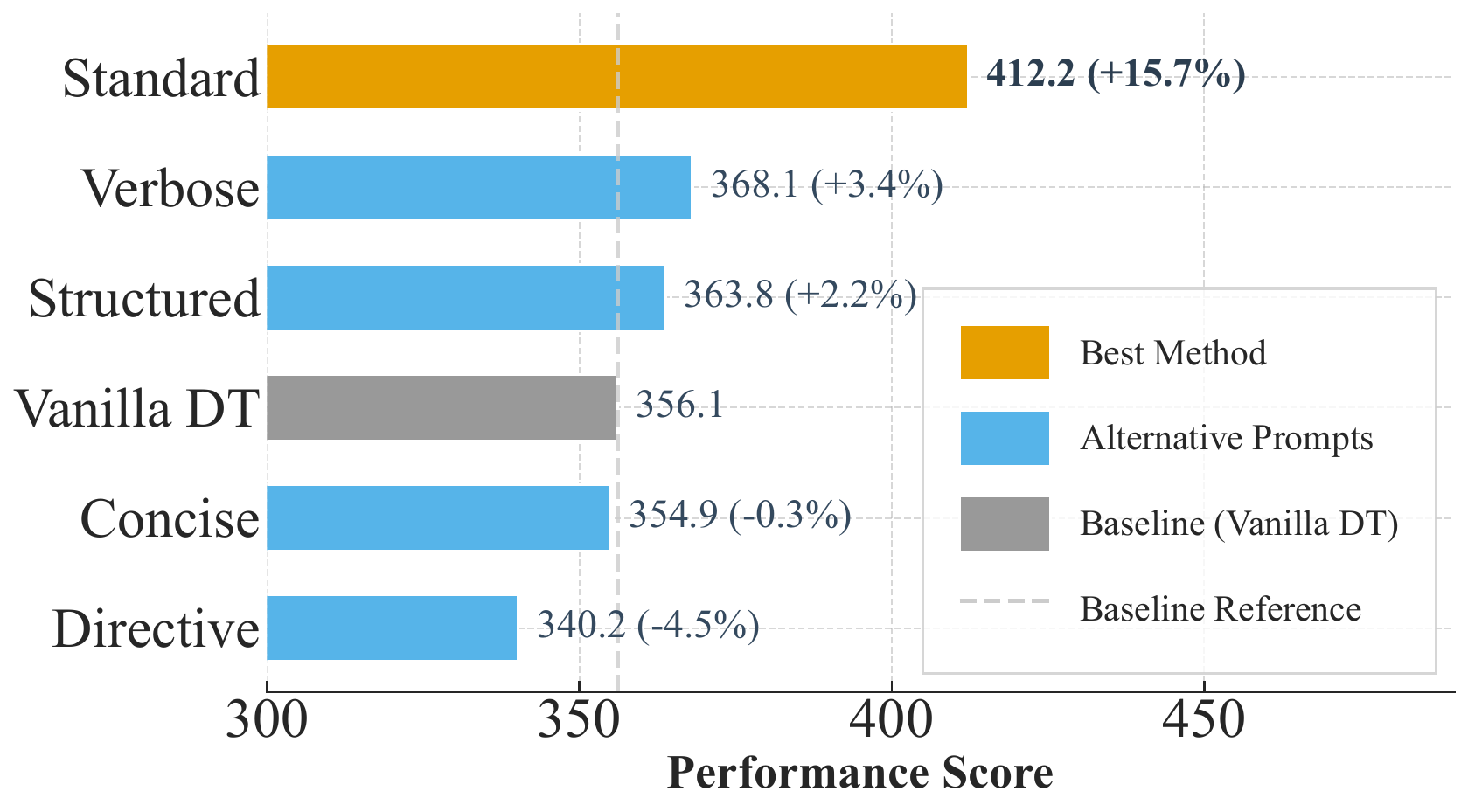}
\caption{Impact of prompt formulation on AuctionNet-High (100\% budget). 
Standard prompts perform best, while overly concise, directive, or verbose prompts degrade performance.
}
\label{fig:prompt-formulation}
\vspace{-3mm}

\end{figure}

Prompt design follows a clear ``Goldilocks'' principle: balanced descriptions outperform both extremes, and the optimal prompt is neither the shortest nor the most detailed. Figure~\ref{fig:prompt-formulation} compares five prompt formulations on AuctionNet-High under the 100\% budget setting.

Directive prompts perform the worst (340.2, $-4.5\%$ vs.\ DT). Pure imperatives (e.g., ``bid higher/lower'') provide little situational grounding, so the encoder cannot form stable, decision-relevant representations beyond generic sentiment. Concise prompts are similarly insufficient (354.9, $-0.3\%$), omitting the context needed to interpret objectives, recent outcomes, and strategy cues. On the other end, Verbose prompts reduce the gain (368.1, $+3.4\%$): excessive detail introduces irrelevant variation that dilutes the control-relevant signal, consistent with the finding that larger LLMs (which produce richer representations) are also not necessarily better (Section~\ref{sec:encoder-sensitivity}).
Structured prompts with explicit field labels offer only a marginal improvement (363.8, $+2.2\%$), indicating that rigid templates do not automatically improve alignment. The best performance comes from Standard prompts ($+15.7\%$), which provide enough situational context (objectives, feedback, guidance) without overwhelming the encoder. This suggests a practical design guideline: \textit{semantic prompts for decision-making should prioritize information density and relevance over linguistic elaboration}.

\begin{table}[h]
\caption{SemBid with different LLM encoders on AuctionNet-High (100\% budget). Imp.\ reports relative improvement over vanilla DT.}
\label{tab:encoder-comparison}
\vspace{-2mm} 

\centering
\begin{small}
\begin{sc}
\resizebox{\columnwidth}{!}{
\begin{tabular}{lccc}
\toprule
Model & Params & Score & Imp. (vs DT) \\
\midrule
Vanilla DT (Baseline) & - & 356.14 & - \\
\midrule
Qwen2.5-Instruct & 0.5B & \textbf{412.20} & \textbf{+15.7\%} \\
Qwen2.5-Instruct & 1.5B & 389.43 & +9.3\% \\
Qwen2.5-Instruct & 7B & 383.81 & +7.8\% \\
\midrule
DeepSeek-R1-Distill & 1.5B & 389.40 & +9.3\% \\
GLM-4-Chat & 9B & 389.03 & +9.2\% \\
\bottomrule
\end{tabular}
}
\end{sc}
\end{small}

\vspace{-5mm}
\end{table}

\subsubsection{Sensitivity to Encoder Variants}
\label{sec:encoder-sensitivity}

Table~\ref{tab:encoder-comparison} reports the performance of SemBid with different LLM encoders on AuctionNet-High under the 100\% budget setting. 
We focus on general-purpose LLMs including the Qwen2.5-Instruct family at multiple scales (0.5B, 1.5B, and 7B)~\cite{hui2024qwen2}, the distilled reasoning model DeepSeek-R1-Distill (1.5B)~\cite{guo2025deepseek}, and the chat-oriented GLM-4-Chat (9B)~\cite{glm2024chatglm}. 

All encoders improve over vanilla DT, but smaller models outperform larger ones---challenging the common assumption that bigger LLMs always produce more useful representations. The smallest model, Qwen2.5-Instruct 0.5B, achieves the best score (412.20, $+15.7\%$), surpassing its larger counterparts at 1.5B ($+9.3\%$) and 7B ($+7.8\%$). Similarly, DeepSeek-R1-Distill (1.5B) and GLM-4-Chat (9B) yield comparable but lower improvements ($\sim$$+9\%$), despite their stronger general reasoning capabilities.

These observations suggest that, for semantic-enhanced decision transformers, representation usefulness is driven more by task alignment than by raw model capacity or general reasoning strength. Larger or more heavily optimized models tend to encode richer and more abstract linguistic features, which may be beneficial for open-ended reasoning but less aligned with the concise, templated semantic prompts used to describe bidding objectives, historic outcomes, and strategy directives.

\section{Conclusion}
This work revisits auto-bidding from the perspective that offline logs encode not only numerical dynamics but also implicit semantics about objectives, constraints, and strategic intent. Our preliminary studies suggest that LLM-derived representations can capture decision-relevant cues beyond raw state features, but that the benefit is not automatic: different semantic dimensions behave differently, and naive prompt/encoder choices can even hurt performance. Motivated by these observations, we propose \textit{SemBid}, which augments a Decision Transformer with three complementary semantic signals---\textit{Task}, \textit{History}, and \textit{Strategy}---to provide objective grounding, contextual feedback, and actionable guidance.Extensive experiments on AuctionNet, spanning varied reward densities and budget regimes, show that this semantic augmentation delivers consistent gains over strong offline RL and generative bidding baselines, with the largest improvements occurring under more generous budget regimes and higher-conversion conditions. Overall, our results indicate that decision transformers offer a natural interface for integrating language priors with numerical trajectories, and that well-designed semantic inputs can serve as a practical lever for improving auto-bidding policies.

\clearpage

\bibliographystyle{ACM-Reference-Format}
\bibliography{sample-base}

\newpage
\appendix
\section{Preliminary Study: Detailed Results}
\label{app:preliminary}

This appendix reports the \textbf{complete} set of preliminary experiments referenced in Section~\ref{sec:preliminary}. In addition to the results summarized in the main paper, Appendix~\ref{app:preliminary} includes \textbf{supplementary analyses} that are omitted from the main text for brevity, such as a broader exploration of prompt variants, prompt grouping into Task/History/Strategy, and the \textbf{cross-category correlation} (CCA) used to assess redundancy and complementarity.

This study validates four aspects: (1) semantic content in LLM embeddings, (2) limitations of using embeddings alone, (3) complementarity across prompt types, and (4) effectiveness of fusion mechanisms.

\subsection{Semantic Content in LLM Embeddings}

We first verify whether LLM embeddings encode bidding-relevant information or are merely random noise. We adopt linear probing: if a simple linear model can predict bidding actions from embeddings, the representations contain task-relevant information.

We first \textbf{render numerical bidding states into natural-language text} using rule-based templates (e.g., ``Budget: 80\%. pValue: High. Time remaining: 2 hours.''), and then \textbf{encode the text} with a frozen Qwen-0.5B to obtain state embeddings. We compare against random baselines. Crucially, we include a shuffled test where \textbf{sample--embedding correspondences are randomized} to remove semantic alignment while preserving the embedding distribution.

\begin{table}[H]
\caption{LLM embeddings vs.\ random baselines for action prediction (linear probing).}
\label{tab:preliminary-linear-app}
\centering
{\small
\setlength{\tabcolsep}{6pt}
\renewcommand{\arraystretch}{1.05}
\begin{tabular}{lcc}
\toprule
Input Type & $R^2$ & vs Random \\
\midrule
Random Gaussian & -0.143 & baseline \\
Shuffled Pairing (Text--Embedding) & -0.015 & - \\
\textbf{State Text Embedding} & \textbf{0.121} & \textbf{+946\%} \\
\bottomrule
\end{tabular}
}
\end{table}

State Text Embedding achieves $R^2=0.121$, substantially outperforming random baselines (+946\% over random Gaussian). Most importantly, randomizing sample--embedding correspondences reduces $R^2$ to -0.015, confirming that \textbf{semantic alignment}, rather than incidental statistics, drives prediction. This establishes that \textbf{LLM embeddings encode bidding-relevant semantics}.

\subsection{Limits of Using LLM Embeddings Alone}

Given that LLM embeddings contain useful semantics, a natural question arises: can we bypass numerical features entirely and bid using only semantic representations?

\begin{table}[H]
\caption{Semantic vs.\ numerical features for bid prediction.}
\label{tab:preliminary-alone-app}
\centering
{\small
\setlength{\tabcolsep}{6pt}
\renewcommand{\arraystretch}{1.05}
\begin{tabular}{lccc}
\toprule
Input Type & $R^2$ & MSE & MAE \\
\midrule
Numerical Features Only & 0.847 & 6.098 & 1.417 \\
LLM Embedding Only & 0.121 & 35.499 & 3.52 \\
Numerical + Embedding & 0.852 & 5.89 & 1.38 \\
\bottomrule
\end{tabular}
}
\end{table}

Results indicate that embeddings alone are insufficient. Numerical features achieve $R^2=0.847$, while semantic embeddings alone reach only 0.121---an 85.7\% performance gap. This is expected: bidding requires fine-grained numerical reasoning (e.g., budget levels, conversion likelihoods, and cost accounting) that cannot be faithfully represented by language alone. Therefore, \textbf{semantic signals should complement, not replace, numerical features}.

\subsection{Complementarity Across Prompt Types}

Having established that embeddings contain useful information but cannot work alone, we investigate whether different prompt formulations capture distinct aspects of the bidding context. We construct a \textbf{representative set of 18 prompt templates} and organize them into three semantic categories---Task, History, and Strategy---to balance coverage and compactness. Appendix~\ref{app:semantic_prompt_templates} lists a broader pool of prompt templates used across ablations; here we focus on the \textbf{core subset} used for this preliminary analysis.

We list the effective prompts ($R^2>0.05$) in Table~\ref{tab:prompt-effective-app}.

\begin{table}[H]
\vspace{-4mm}

\caption{Effective prompts with $R^2>0.05$.}
\label{tab:prompt-effective-app}
\centering
{\small
\setlength{\tabcolsep}{6pt}
\renewcommand{\arraystretch}{1.05}
\begin{tabular}{llc}
\toprule
Category & Prompt & $R^2$ \\
\midrule
Task & task\_roi\_objective & 0.152 \\
Task & task\_budget\_constraint & 0.132 \\
Task & task\_combined\_metrics & 0.132 \\
Task & task\_efficiency\_ratio & 0.130 \\
History & history\_progress\_summary & 0.132 \\
History & history\_budget\_status & 0.132 \\
History & history\_time\_progress & 0.112 \\
History & history\_spend\_rate & 0.112 \\
History & history\_time\_remaining & 0.054 \\
Strategy & strategy\_conversion\_rate & 0.152 \\
Strategy & strategy\_optimization\_signal & 0.148 \\
Strategy & strategy\_time\_pressure & 0.132 \\
Strategy & strategy\_budget\_pacing & 0.112 \\
\bottomrule
\end{tabular}
}
\vspace{-3mm}

\end{table}

We quantify redundancy and complementarity using canonical correlation analysis (CCA) between prompt embeddings. The overall average correlation is 0.123, suggesting low redundancy in general; however, History--Strategy shows higher correlation (0.201), indicating partial overlap (Table~\ref{tab:prompt-cca-app}).

\begin{table}[H]
\caption{Embedding correlation across prompt categories measured by CCA.}
\label{tab:prompt-cca-app}
\centering
{\small
\setlength{\tabcolsep}{6pt}
\renewcommand{\arraystretch}{1.05}
\begin{tabular}{lc}
\toprule
Pair & Avg CCA \\
\midrule
Task -- History & 0.108 \\
Task -- Strategy & 0.059 \\
History -- Strategy & 0.201 \\
\bottomrule
\end{tabular}
}
\vspace{-3mm}

\end{table}

Finally, we evaluate each category alone and in combination using linear probing (Table~\ref{tab:preliminary-complementarity-app}). Combining all three yields $R^2=0.204$, a 52.26\% improvement over the best single category. The incremental gains (Strategy +0.072, History +0.055, Task +0.005 in $\Delta R^2$) further support complementarity.

\begin{table}[H]
\vspace{-3mm}
\caption{Single vs.\ combined prompt categories (linear probing).}
\label{tab:preliminary-complementarity-app}
\centering
{\small
\setlength{\tabcolsep}{6pt}
\renewcommand{\arraystretch}{1.05}
\begin{tabular}{lccc}
\toprule
Configuration & $R^2$ & MSE & MAE \\
\midrule
Task only & 0.142 & 40.07 & 4.03 \\
History only & 0.121 & 36.02 & 4.05 \\
Strategy only & 0.134 & 35.49 & 3.55 \\
Task + History & 0.132 & 35.58 & 3.94 \\
Task + Strategy & 0.149 & 34.86 & 3.57 \\
History + Strategy & 0.199 & 32.82 & 3.54 \\
\textbf{Task + History + Strategy} & \textbf{0.204} & \textbf{32.63} & \textbf{3.67} \\
\bottomrule
\end{tabular}
}
\vspace{-3mm}
\end{table}

Task embeddings perform best individually ($R^2=0.142$), followed by Strategy ($R^2=0.134$) and History ($R^2=0.121$). This superadditive effect indicates that \textbf{different prompt types encode complementary information}: Task prompts capture objective grounding, History prompts summarize short-term feedback, and Strategy prompts provide domain heuristics.

\subsection{Fusion Mechanisms for Semantic Signals}

Finally, we investigate how to integrate semantic embeddings with numerical features. We compare simple concatenation against more structured fusion mechanisms.

\begin{table}[H]
\vspace{-3mm}
\caption{Fusion mechanism comparison.}
\label{tab:preliminary-fusion-app}
\centering
{\small
\setlength{\tabcolsep}{6pt}
\renewcommand{\arraystretch}{1.05}
\begin{tabular}{lcc}
\toprule
Fusion Method & $R^2$ & vs Baseline \\
\midrule
Numerical Only (Baseline) & 0.880 & - \\
\midrule
Concat Fusion & 0.850 & -3.41\% \\
Residual Fusion & 0.872 & -0.91\% \\
Gated Fusion & 0.863 & -1.93\% \\
\textbf{Cross-Attention Fusion} & \textbf{0.886} & \textbf{+0.68\%} \\
FiLM Fusion & 0.859 & -2.39\% \\
\bottomrule
\end{tabular}
}
\vspace{-3mm}
\end{table}

Simple concatenation performs worst, degrading performance by 3.41\%. This suggests that treating embeddings as uniformly informative can dilute the strong numerical signal. In contrast, \textbf{cross-attention} yields the best result and provides a consistent improvement over the numerical baseline, indicating that attention-based fusion can selectively filter and integrate semantic context. Other lightweight fusion variants (residual/gated/FiLM) remain slightly below the numerical baseline in this controlled setting, motivating our attention-based integration in the full SemBid architecture.

\subsection{Key Findings}

These preliminary experiments (Section~\ref{sec:preliminary} and Appendix~\ref{app:preliminary}) provide concrete evidence supporting SemBid's design:
\begin{itemize}[leftmargin=*, itemsep=0pt, topsep=2pt, parsep=0pt]
\item LLM embeddings encode bidding semantics (+946\% over random; shuffling correspondence removes predictive power).
\item Embeddings alone are insufficient (85.7\% gap vs numerical features).
\item Prompt types exhibit complementarity (Task/History/Strategy combined yields +52.3\% over the best single category).
\item Proper fusion matters: cross-attention improves over naive concatenation and yields the strongest performance in this study.
\end{itemize}

Overall, the results justify a \textbf{compositional semantic design} (multiple prompt types) and a \textbf{selective integration mechanism} (attention) as the basis for SemBid.

\section{Semantic Signal Generation Details}
\label{app:semantic_signals}

This section details the prompt generation logic for the three semantic signals used in SemBid. The generation process is grounded in a rule-based template system designed to align numerical state features with the LLM's pre-training corpus. Unless otherwise stated, the rules below correspond to the \emph{high-conversion} setting. For the low-conversion setting, we keep the same semantic categories and underlying state variables, but render feedback in a sparsity-aware manner (e.g., conversion happened vs.\ no conversion) and adjust scale-dependent thresholds in strategy rules when applicable (see Appendix~\ref{app:semantic_prompt_templates}).

\subsection{Signal 1: Task Description (Goal Semantics)}
\label{app:task_signal_logic}

The task signal $T_t$ provides a natural language specification of the advertiser's optimization goal. To prevent the model from overfitting to a fixed sentence structure, we employ a \textbf{stochastic template selection} mechanism.

\textbf{Generation Logic:}
At each timestep, the system randomly samples a template from a pre-defined set $\mathcal{P}_{task}$ and populates it with the specific Target CPA ($C_{target}$). This injects linguistic diversity while maintaining semantic consistency. The formal process is described in Algorithm~\ref{alg:task_signal}.

\begin{algorithm}[h]
\caption{Task Signal Generation via Stochastic Sampling}
\label{alg:task_signal}
\begin{algorithmic}[1]
\REQUIRE Target CPA $C_{target}$
\STATE \textbf{Define Template Set} $\mathcal{P}_{task} = \{$
\STATE \quad ``Maximize conversions while maintaining CPA below \{cpa:.1f\}.'',
\STATE \quad ``Optimize bidding to achieve target CPA of \{cpa:.1f\}.'',
\STATE \quad ``Control cost per acquisition to stay within \{cpa:.1f\}.'',
\STATE \quad ``Balance conversion volume and cost efficiency with target CPA \{cpa:.1f\}.'' $\}$
\STATE $T_{template} \leftarrow \text{RandomSample}(\mathcal{P}_{task})$
\STATE $T_t \leftarrow \text{Format}(T_{template}, \text{cpa}=C_{target})$
\RETURN $T_t$
\end{algorithmic}
\end{algorithm}

\subsection{Signal 2: History }
\label{app:history_signal_logic}

The history signal $H_t$ translates the immediate feedback from the previous action into qualitative assessments. The system monitors three key performance indicators: ROI, conversion trend (proxied by change in conversion rate), and cost efficiency.

\textbf{Metric Discretization Rules:}
We define threshold-based rules to map numerical changes to semantic descriptions.
Let $s_{t-1}, a_{t-1}, s_t$ denote the transition tuple, and let $\Delta \text{CVR}$ and $\Delta \text{CPA}$ denote changes in conversion rate and CPA, respectively.

\begin{enumerate}
    \item \textbf{ROI Assessment:}
    The approximate ROI is calculated with a conversion weight factor $\omega=10.0$.
    We add a small constant $\epsilon$ to avoid division by zero:
    \begin{equation}
        \text{ROI} \approx \frac{\omega \cdot \Delta \text{Conv} - \Delta \text{Cost}}{\Delta \text{Cost} + \epsilon}
    \end{equation}
    The semantic mapping is defined as:
    \begin{itemize}
        \item If $\text{ROI} < 0.5$: $\to$ ``The ROI was low.''
        \item If $0.5 \le \text{ROI} < 1.5$: $\to$ ``The ROI was moderate.''
        \item If $\text{ROI} \ge 1.5$: $\to$ ``The ROI was good.''
    \end{itemize}
    
    \item \textbf{Conversion Trend (CVR-based):}
    \begin{itemize}
        \item If $\Delta \text{CVR} > 0.001$: $\to$ ``Conversion rate increased.''
        \item If $\Delta \text{CVR} < -0.001$: $\to$ ``Conversion rate decreased.''
    \end{itemize}
    
    \item \textbf{Cost Efficiency (CPA):}
    Let $\Delta \text{CPA} = \text{CPA}_t - \text{CPA}_{t-1}$.
    \begin{itemize}
        \item If $\Delta \text{CPA} > \tau_{\text{cpa}}$: $\to$ ``CPA rose.''
        \item If $\Delta \text{CPA} < -\tau_{\text{cpa}}$: $\to$ ``CPA dropped.''
    \end{itemize}
\end{enumerate}

\subsection{Signal 3: Strategy (Foresight)}
\label{app:strategy_signal_logic}

The strategy signal $S_t$ provides actionable advice derived from a \textbf{composite heuristic analysis}. The system sequentially analyzes multiple state factors (conversion probability, budget status, and action magnitude) and concatenates relevant advice strings.

\textbf{Heuristic Components:}

\paragraph{1. Value Analysis (pValue):}
Evaluates the conversion probability $p$ of the current impression batch.
\begin{itemize}
\item \textbf{High Opportunity} ($p > 0.01$): Suggests ``High pValue indicates good opportunity.''
\item \textbf{Low Opportunity} ($p < 0.001$): Suggests ``Low pValue suggests lower conversion potential.''
\end{itemize}

\paragraph{2. Budget Health:}
Evaluates the remaining budget ratio $R_b = B_{left} / B_{total}$.
\begin{itemize}
\item \textbf{Scarcity} ($R_b < 0.2$): ``Remaining budget is low. Bid conservatively.''
\item \textbf{Abundance} ($R_b > 0.7$): ``Remaining budget is sufficient. You can bid more aggressively.''
\end{itemize}

\paragraph{3. Reference-based Guidance:}
To facilitate stable learning, we provide bidding suggestions based on the magnitude of the suggested bid $b_t$.
\begin{itemize}
\item If $b_t > 50$: ``Consider increasing the bid...'' (Aggressive)
\item If $b_t < 10$: ``Consider bidding conservatively...'' (Conservative)
\item Otherwise: ``Consider a balanced bidding approach.'' (Moderate)
\end{itemize}
\textit{Note: These thresholds are specific to the high-conversion setting described above. For the low-conversion subset, we use scale-adjusted bid thresholds to account for the larger action range.}
\section{Dataset and Environment Details}
\label{app:dataset}

\subsection{Dataset Composition and Market Scenarios}
To rigorously evaluate the model's adaptability to diverse market conditions, we curated three distinct experimental subsets from the AuctionNet log data. These subsets represent varying levels of conversion difficulty, constraint strictness, and bidding aggression. The detailed statistics are summarized in Table~\ref{tab:dataset_stats}.

\begin{table}[htbp]
\centering
\caption{Parameters Statistics per Conversion Level}
\label{tab:dataset_stats}

\resizebox{\linewidth}{!}{
    \begin{tabular}{cccc} 
    \toprule
    \textbf{Params} & \textbf{Low conversion} & \textbf{Medium conversion} & \textbf{High conversion} \\
    \midrule
    Trajectories & 17,773 & 17,257 & 17,257 \\
    Delivery Periods & 8,371 & 8,253 & 8,253 \\
    Time steps in a trajectory & 48 & 48 & 48 \\
    State dimension & 16 & 16 & 16 \\
    Action dimension & 1 & 1 & 1 \\
    Return-To-Go Dimension & 1 & 1 & 1 \\
    Action range & [0.0, 2960.53] & [0.0, 145.47] & [0.0, 145.47] \\
    Impression value range & [0, 1] & [0, 1] & [0, 1] \\
    CPA range & [60.0, 130.0] & [15.0, 50.0] & [6.0, 12.0] \\
    Total conversion range & [0.0, 96.0] & [0, 907.5368] & [0.0, 1475.0] \\
    \bottomrule
    \end{tabular}
}
\end{table}
\subsection{Environment Characterization}
Each dataset poses unique challenges for the auto-bidding agent:

\paragraph{Scenario 1: High Conversion (The Constraint Challenge)}
\begin{itemize}
\item \textbf{Characteristics:} This environment features the highest conversion rate (11.21\%) but imposes the strictest CPA constraints (Target CPA $\approx$ 8.27).
\item \textbf{Challenge:} The primary difficulty is \textbf{Constraint Satisfaction}. The logged data shows a low CPA compliance rate of only 29.17\%, indicating that even human/rule-based bidders struggle to balance high conversion volume with strict cost controls. The agent must learn to bid conservatively despite high pValues.
\end{itemize}

\paragraph{Scenario 2: Medium Conversion (The Budget Challenge)}
\begin{itemize}
\item \textbf{Characteristics:} Represents a standard market environment with moderate CVR (3.85
\item \textbf{Challenge:} The key issue is \textbf{Budget Management}. The historical data exhibits significant over-spending (Budget Usage $\approx$ 123.56\%), requiring the agent to learn superior pacing strategies to avoid premature budget exhaustion.
\end{itemize}

\paragraph{Scenario 3: Low Conversion (The Exploration \& Generalization Challenge)}
\begin{itemize}
    \item \textbf{Characteristics:} This is a ``few-shot'' scenario with only 3 advertisers and sparse rewards (CVR $\approx$ 1.45\%).
    \item \textbf{Challenge:}
    \begin{enumerate}
        \item \textbf{Action Space Shift:} The average bid action ($\mu_a \approx 81.20$) is an order of magnitude larger than in other scenarios ($\mu_a \approx 8.0$). This drastic distribution shift necessitates robust state/action normalization (Z-score) to prevent transfer learning failure.
        \item \textbf{Sparsity:} The extremely low conversion rate tests the agent's ability to perform efficient exploration in sparse-reward environments.
    \end{enumerate}
\end{itemize}

\subsection{Experimental Subsets Configuration}
Based on the analysis above, we configured our experiments as follows:
\begin{enumerate}
    \item \textbf{Sampled Dataset (0.5GB):} Drawn primarily from the High and Low conversion scenarios to test the model's capability in both constraint satisfaction and domain generalization. Used for SemBid and most baselines (BCQ, IQL, GAVE).
    \item \textbf{Full High-Conversion Dataset:} Used specifically for \textbf{GAS} to reproduce its SOTA performance on its native large-scale setting.
    \item \textbf{Full RL Dataset (302GB):} Used for \textbf{Vanilla DT} to evaluate scalability.
\end{enumerate}

\subsection{State Feature Engineering}
\label{app:state-features}
The numerical state $s_t \in \mathbb{R}^{16}$ consists of four logical groups. All features are z-score normalized using statistics computed from the training set.

\begin{enumerate}
    \item \textbf{Temporal (2-dim):} Time remaining ratio, Time elapsed.
    \item \textbf{Budget (4-dim):} Budget remaining ratio, Consumption rate, Budget Pacing, Exhaustion risk.
    \item \textbf{Performance (4-dim):} Current CPA, Target CPA, CPA violation degree, Historical CVR.
    \item \textbf{History (4-dim):} Avg. bid (last 3 steps), Win rate (last 3 steps), Avg. cost (last 3 steps), Bid volatility.
\end{enumerate}

\section{Implementation Details}
\label{app:implementation}

\subsection{Model Architecture and Hyperparameters}
We adopt a Language-Guided Decision Transformer. Unlike standard implementations, we utilize high-dimensional semantic embeddings to preserve the rich information from the LLM.

\textbf{Semantic Embedding Upscaling:}
The base embeddings from Qwen-0.5B ($d=896$) are projected to a higher dimension ($d=2048$) using a randomized linear projection layer during preprocessing. These pre-computed embeddings are then fed into the Decision Transformer.

Complete hyperparameters are listed in Table~\ref{tab:hyperparams}.

\begin{table}[h]
\centering

\resizebox{\linewidth}{!}{
    \begin{tabular}{@{}llc@{}}
    \toprule
    \textbf{Category} & \textbf{Parameter} & \textbf{Value} \\
    \midrule
    \multirow{6}{*}{\textbf{Architecture}} 
     & Transformer Layers & 6 \\
     & Attention Heads & 4 \\
     & Hidden Dimension ($d_{model}$) & 128 \\
     & Feedforward Dimension & 512 \\
     & Activation Function & GELU \\
     & Dropout Rate & 0.1 \\
    \midrule
    \multirow{3}{*}{\textbf{Semantic Encoder}} 
     & LLM Backbone & Qwen-0.5B (Frozen) \\
     & Projection Type & Random Linear ($896 \to 2048$) \\
     & Input Embedding Dim & 2048 \\
    \midrule
    \multirow{6}{*}{\textbf{Training}} 
     & Optimizer & AdamW \\
     & Learning Rate & $1 \times 10^{-4}$ \\
     & Weight Decay & $1 \times 10^{-4}$ \\
     & Batch Size & 64 \\
     & Training Steps & 800,000 \\
     & Max Episode Length & 48 \\
    \bottomrule
    \end{tabular}
}
\caption{Hyperparameters for SemBid and Training Configuration. Note the use of high-dimensional semantic embeddings.}
\label{tab:hyperparams}
\vspace{-4mm}
\end{table}

\subsection{Computational Resources}
All experiments were conducted on a high-performance computing cluster. 
\begin{itemize}
\item \textbf{Hardware:} 1 $\times$ NVIDIA A100 (40GB) GPU.
\item \textbf{Preprocessing:} Offline embedding generation with Qwen-0.5B took $\sim$2 hours.
\item \textbf{Training:} The model was trained for 800k steps with persistent data workers.
\end{itemize}

\section{Baseline Implementation Details}
\label{app:baselines}

To ensure fair comparison, we followed official implementations or standard protocols for all baselines. We standardized checkpointing, model selection, and training horizons across methods. Table~\ref{tab:baseline_configs} summarizes the configurations.

\paragraph{Checkpointing and model selection.}
We saved checkpoints every 10k steps and ran coarse evaluation every 100k steps (5 seeds). We then re-scanned checkpoints within $\pm$100k of the best coarse point at 10k intervals (5 seeds), and reported the best averaged score.

\paragraph{Convergence and implementation notes.}
Most baselines converge within 400k steps; however, DT-style sequence models (\textit{SemBid}, DT, GAVE) can stabilize more slowly in a few runs. To avoid under-training and keep comparisons uniform, we trained all methods for 800k steps. GAS follows the official two-stage pipeline (policy then critic); GAVE uses a larger backbone for value guidance; PID adjusts gains dynamically based on the budget ratio.

\begin{table}[h]
\centering
\caption{Detailed training configurations. Steps indicates the maximum training steps.}
\label{tab:baseline_configs}
\resizebox{\linewidth}{!}{
\begin{tabular}{@{}lcccc@{}}
\toprule
\textbf{Method} & \textbf{Type} & \textbf{Max Steps} & \textbf{Batch} & \textbf{Key Configuration} \\
\midrule
\textbf{SemBid} (Ours) & DT+LLM & 800,000 & 64 & Emb Dim=2048, Random Proj \\
\textbf{GAS}~\cite{li2025gas} & DT+Search & 800,000 & 64 & Two-stage (Policy+Critic), $w=0.2$ \\
\textbf{GAVE}~\cite{gao2025generative} & DT+Value & 800,000 & 64 & Large Arch (L=8, H=512), $\tau=0.99$ \\
\textbf{DT}~\cite{chen2021decision} & Transformer & 800,000 & 64 & Vanilla, Score-based RTG \\
\textbf{TD3+BC}~\cite{fujimoto2021minimalist} & Offline RL & 800,000 & 100 & Standard Implementation \\
\textbf{CQL}~\cite{kumar2020conservative} & Offline RL & 800,000 & 100 & Conservative $\alpha=5.0$ \\
\textbf{BCQ}~\cite{fujimoto2019off} & Offline RL & 800,000 & 100 & Early Convergence \\
\textbf{IQL}~\cite{kostrikov2021offline} & Offline RL & 800,000 & 100 & Implicit Q-Learning \\
\textbf{BC}~\cite{torabi2018behavioral} & Imitation & 800,000 & 100 & Behavior Cloning \\
\bottomrule
\end{tabular}
}
\vspace{-2mm}
\end{table}

\section{Prompt Templates for Semantic Signals}
\label{app:semantic_prompt_templates}

This appendix is organized into two subsections. The first subsection lists the \emph{standard} prompt templates used in our experiments for high- and low-conversion regimes. The second subsection reports the \emph{prompt-variant} templates used only in the high-conversion prompt-design study (RQ3; Section~\ref{sec:preliminary}). We group templates into Task (objective), History (feedback), and Strategy (guidance). The medium-conversion setting reuses the high-conversion pool.

\subsection{Standard Templates for High- and Low-Conversion Settings}
\label{app:prompt_standard}
Placeholders follow Python-style formatting (e.g., \{cpa:.1f\}) used in our preprocessing scripts.

\paragraph{High-Conversion Templates.}
Figure~\ref{fig:prompt_semantic_high} lists the high-conversion template pool used in our experiments.
\paragraph{Low-Conversion Templates.}
The Task templates are shared across regimes; differences mainly appear in History feedback and Strategy guidance due to sparsity and scale shifts. Due to extreme reward sparsity in the low-conversion subset, we render feedback using event-based conversion indicators instead of short-term CVR trend descriptors. Figure~\ref{fig:prompt_semantic_low} lists the low-conversion template pool used in our experiments.

\subsection{Prompt Variants }
\label{app:prompt_variants_high}

This subsection lists the alternative prompt styles used only in the high-conversion prompt-variant study. We do not vary prompt styles in the low-conversion subset to avoid confounding prompt sensitivity with sparsity-driven rendering changes.

\begin{figure*}[th!]
\begin{AIBox}[title={Prompt Templates (High-Conversion)}]
\parbox[t]{\textwidth}{
\small

\PromptSection{Task Description (4 variants)}
\begin{enumerate}
\item Balance conversion volume and cost efficiency with target CPA \{cpa:.1f\}.
\item Maximize conversions while maintaining CPA below \{cpa:.1f\}.
\item Optimize bidding to achieve target CPA of \{cpa:.1f\}.
\item Control cost per acquisition to stay within \{cpa:.1f\}.
\end{enumerate}

\noindent\textcolor{AIGrayLine}{\rule{\linewidth}{0.35pt}}

\PromptSection{History (7 categories $\times$ 5 variants)}
\begin{itemize}
\item \textbf{ROI Low}: The ROI was low after the last bid; Previous bid resulted in low return on investment; Last action led to poor ROI performance; The bid was not profitable; ROI dropped below expectations.
\item \textbf{ROI Moderate}: The ROI was moderate after the last bid; Previous bid achieved acceptable ROI; Last action resulted in moderate returns; The bid showed moderate profitability; ROI was within acceptable range.
\item \textbf{ROI Good}: The ROI was good after the last bid; Previous bid achieved high return on investment; Last action resulted in excellent ROI; The bid was highly profitable; ROI exceeded expectations.
\item \textbf{CVR Increase}: Conversion rate increased after the bid; CVR improved following the action; The bid led to higher conversion rate; Conversion performance improved; CVR showed positive trend.
\item \textbf{CVR Decrease}: Conversion rate decreased after the bid; CVR dropped following the action; The bid led to lower conversion rate; Conversion performance declined; CVR showed negative trend.
\item \textbf{CPA Increase}: Cost per acquisition increased; CPA rose after the bid; Acquisition cost went up; CPA showed upward trend; Cost efficiency decreased.
\item \textbf{CPA Decrease}: Cost per acquisition decreased; CPA dropped after the bid; Acquisition cost went down; CPA showed downward trend; Cost efficiency improved.
\end{itemize}

\noindent\textcolor{AIGrayLine}{\rule{\linewidth}{0.35pt}}

\PromptSection{Strategy (7 categories $\times$ 5 variants)}
\begin{itemize}
\item \textbf{Conservative}: Consider bidding conservatively to maintain budget; Use a conservative bid to preserve resources; A moderate bid would be safer now; Keep the bid conservative to avoid overspending; Maintain conservative bidding strategy.
\item \textbf{Moderate}: A moderate bid would be appropriate; Consider a balanced bidding approach; Use a moderate bid to balance risk and reward; Moderate bid would help maintain stability; Try a moderate bidding strategy.
\item \textbf{Aggressive}: Consider increasing the bid to capture more opportunities; An aggressive bid might capture better conversions; Raise the bid to compete more effectively; Higher bid could improve win rate; Consider bidding more aggressively.
\item \textbf{High pValue}: High pValue suggests good conversion potential; The pValue is favorable, you can bid higher; Strong pValue indicates good opportunity; High pValue means better conversion probability; Favorable pValue supports higher bidding.
\item \textbf{Low pValue}: Low pValue suggests lower conversion potential; The pValue is low, bid conservatively; Weak pValue indicates limited opportunity; Low pValue means lower conversion probability; Unfavorable pValue suggests lower bidding.
\item \textbf{Budget Low}: Remaining budget is low; Limited budget remaining, use cautious bidding; Low budget requires careful bid management; Conserve budget with moderate bids; Budget constraint suggests conservative approach.
\item \textbf{Budget High}: Remaining budget is sufficient; Adequate budget allows for higher bids; Good budget availability supports competitive bidding; Sufficient budget enables flexible bidding strategy; Budget is healthy, consider more aggressive bids.
\end{itemize}
}
\end{AIBox}
\caption{High-conversion template pool used in our experiments.}
\label{fig:prompt_semantic_high}
\end{figure*}

\begin{figure*}[th!]
\begin{AIBox}[title={Prompt Templates (Low-Conversion)}]
\parbox[t]{\textwidth}{
\small

\PromptSection{Task Description (4 variants)}
\begin{enumerate}
\item Balance conversion volume and cost efficiency with target CPA \{cpa:.1f\}.
\item Maximize conversions while maintaining CPA below \{cpa:.1f\}.
\item Optimize bidding to achieve target CPA of \{cpa:.1f\}.
\item Control cost per acquisition to stay within \{cpa:.1f\}.
\end{enumerate}

\noindent\textcolor{AIGrayLine}{\rule{\linewidth}{0.35pt}}

\PromptSection{History (7 categories $\times$ 5 variants)}
\begin{itemize}
\item \textbf{ROI Low}: The ROI was low after the last bid; Previous bid resulted in low return on investment; Last action led to poor ROI performance; The bid was not profitable; ROI dropped below expectations.
\item \textbf{ROI Moderate}: The ROI was moderate after the last bid; Previous bid achieved acceptable ROI; Last action resulted in moderate returns; The bid showed moderate profitability; ROI was within acceptable range.
\item \textbf{ROI Good}: The ROI was good after the last bid; Previous bid achieved high return on investment; Last action resulted in excellent ROI; The bid was highly profitable; ROI exceeded expectations.
\item \textbf{Conversion Happened}: A conversion happened after the bid; Conversion was observed following the action; The bid led to a conversion; Conversion occurred after this bid; A conversion was achieved.
\item \textbf{No Conversion}: No conversion was observed after the bid; No conversion happened following the action; The bid did not lead to conversion; No conversion was recorded; Conversion did not occur this time.
\item \textbf{CPA Increase}: Cost per acquisition increased; CPA rose after the bid; Acquisition cost went up; CPA showed upward trend; Cost efficiency decreased.
\item \textbf{CPA Decrease}: Cost per acquisition decreased; CPA dropped after the bid; Acquisition cost went down; CPA showed downward trend; Cost efficiency improved.
\end{itemize}

\noindent\textcolor{AIGrayLine}{\rule{\linewidth}{0.35pt}}

\PromptSection{Strategy (9 categories $\times$ 5 variants)}
\begin{itemize}
\item \textbf{Conservative}: Consider bidding conservatively to maintain budget; Use a conservative bid to preserve resources; A moderate bid would be safer now; Keep the bid conservative to avoid overspending; Maintain conservative bidding strategy.
\item \textbf{Moderate}: A moderate bid would be appropriate; Consider a balanced bidding approach; Use a moderate bid to balance risk and reward; Moderate bid would help maintain stability; Try a moderate bidding strategy.
\item \textbf{Aggressive}: Consider increasing the bid to capture more opportunities; An aggressive bid might capture better conversions; Raise the bid to compete more effectively; Higher bid could improve win rate; Consider bidding more aggressively.
\item \textbf{High pValue}: High pValue suggests good conversion potential. Consider higher bid; The pValue is favorable, you can bid higher; Strong pValue indicates good opportunity; High pValue means better conversion probability; Favorable pValue supports higher bidding.
\item \textbf{Mid pValue}: Mid pValue suggests stable conversion potential; The pValue is moderate, keep bids balanced; Average pValue supports a steady bid; pValue is moderate, avoid extremes; Moderate pValue indicates cautious optimism.
\item \textbf{Low pValue}: Low pValue suggests lower conversion potential. Consider conservative bid; The pValue is low, bid conservatively; Weak pValue indicates limited opportunity; Low pValue means lower conversion probability; Unfavorable pValue suggests lower bidding.
\item \textbf{Budget Low}: Remaining budget is low. Bid conservatively; Limited budget remaining, use cautious bidding; Low budget requires careful bid management; Conserve budget with moderate bids; Budget constraint suggests conservative approach.
\item \textbf{Budget Mid}: Budget is moderate; keep bids balanced; Adequate budget supports steady bidding; Budget level allows for moderate bids; Maintain budget discipline with balanced bids; Moderate budget supports stable bidding.
\item \textbf{Budget High}: Remaining budget is sufficient. You can bid more aggressively; Adequate budget allows for higher bids; Good budget availability supports competitive bidding; Sufficient budget enables flexible bidding strategy; Budget is healthy, consider more aggressive bids.
\end{itemize}
}
\end{AIBox}
\caption{Low-conversion template pool used in our experiments.}
\label{fig:prompt_semantic_low}
\end{figure*}

\begin{figure*}[th!]
\begin{AIBox}[title={Prompt Templates: Directive Style}]
\parbox[t]{\textwidth}{
\small
\PromptSection{Task Description (4 variants)}
\begin{enumerate}
\item Target CPA: \{cpa\}. Optimize bids.
\item Achieve CPA \{cpa\}. Adjust bidding accordingly.
\item Maintain CPA at \{cpa\}. Bid strategically.
\item CPA goal \{cpa\}. Execute optimal bids.
\end{enumerate}
\noindent\textcolor{AIGrayLine}{\rule{\linewidth}{0.35pt}}
\PromptSection{History (7 categories $\times$ 5 variants)}
\begin{itemize}
\item \textbf{ROI Low}: ROI decreased. Adjust strategy; Low ROI observed. Recalibrate approach; ROI fell. Modify bidding; Poor ROI recorded. Change tactics; ROI dropped. Revise strategy.
\item \textbf{ROI Moderate}: ROI moderate. Maintain approach; Acceptable ROI observed. Continue current strategy; ROI within range. Sustain tactics; Moderate ROI recorded. Proceed as is; ROI acceptable. Keep strategy.
\item \textbf{ROI Good}: ROI high. Leverage success; Strong ROI observed. Continue effective approach; ROI exceeded target. Maintain tactics; Excellent ROI recorded. Sustain strategy; ROI strong. Keep approach.
\item \textbf{CVR Increase}: CVR increased. Note improvement; Conversion rate up. Acknowledge trend; CVR rose. Track progress; Conversion improved. Monitor continuation; CVR higher. Observe pattern.
\item \textbf{CVR Decrease}: CVR decreased. Note decline; Conversion rate down. Acknowledge trend; CVR fell. Track regression; Conversion declined. Monitor situation; CVR lower. Observe pattern.
\item \textbf{CPA Increase}: CPA increased. Control costs; Acquisition cost up. Manage spending; CPA rose. Tighten efficiency; Cost increased. Adjust bids; CPA higher. Reduce spending.
\item \textbf{CPA Decrease}: CPA decreased. Maintain efficiency; Acquisition cost down. Sustain control; CPA fell. Continue approach; Cost decreased. Keep strategy; CPA lower. Maintain tactics.
\end{itemize}
\noindent\textcolor{AIGrayLine}{\rule{\linewidth}{0.35pt}}
\PromptSection{Strategy (7 categories $\times$ 5 variants)}
\begin{itemize}
\item \textbf{Conservative}: Bid conservatively. Manage risk; Use low bids. Protect resources; Employ cautious bidding. Control costs; Bid low. Preserve budget; Apply conservative strategy. Minimize exposure.
\item \textbf{Moderate}: Bid moderately. Balance factors; Use balanced bids. Optimize trade-offs; Employ moderate bidding. Maintain equilibrium; Bid mid-range. Balance objectives; Apply moderate strategy. Sustain balance.
\item \textbf{Aggressive}: Bid higher. Capture opportunities; Use competitive bids. Maximize reach; Employ aggressive bidding. Increase coverage; Bid high. Compete effectively; Apply aggressive strategy. Expand presence.
\item \textbf{High pValue}: pValue \{pvalue\}. Strong signal detected; Signal quality high at \{pvalue\}. Favorable conditions; pValue \{pvalue\}. Good opportunity indicated; Strong signal: \{pvalue\}. Positive assessment; pValue \{pvalue\}. High-quality signal present.
\item \textbf{Low pValue}: pValue \{pvalue\}. Weak signal detected; Signal quality low at \{pvalue\}. Unfavorable conditions; pValue \{pvalue\}. Limited opportunity indicated; Weak signal: \{pvalue\}. Negative assessment; pValue \{pvalue\}. Low-quality signal present.
\item \textbf{Budget Low}: Budget at \{budget\}. Conserve resources; Remaining: \{budget\}. Manage carefully; Budget \{budget\}. Control spending; Available: \{budget\}. Preserve allocation; Budget position \{budget\}. Extend runway.
\item \textbf{Budget High}: Budget at \{budget\}. Leverage resources; Remaining: \{budget\}. Utilize availability; Budget \{budget\}. Deploy strategically; Available: \{budget\}. Maximize allocation; Budget position \{budget\}. Capitalize on health.
\end{itemize}
}
\end{AIBox}
\caption{Prompt templates for Directive style (high-conversion prompt-variant study): imperative commands with minimal explanation.}
\label{fig:prompt_directive}
\end{figure*}

\begin{figure*}[th!]
\begin{AIBox}[title={Prompt Templates: Concise Style}]
\parbox[t]{\textwidth}{
\small
\PromptSection{Task Description (4 variants)}
\begin{enumerate}
\item CPA target: \{cpa\}.
\item Target CPA \{cpa\}.
\item CPA goal \{cpa\}.
\item Aim for CPA \{cpa\}.
\end{enumerate}
\noindent\textcolor{AIGrayLine}{\rule{\linewidth}{0.35pt}}
\PromptSection{History (7 categories $\times$ 5 variants)}
\begin{itemize}
\item \textbf{ROI Low}: ROI fell; ROI low; ROI dropped; ROI poor; ROI decreased.
\item \textbf{ROI Moderate}: ROI moderate; ROI acceptable; ROI decent; ROI okay; ROI normal.
\item \textbf{ROI Good}: ROI high; ROI strong; ROI excellent; ROI good; ROI rose.
\item \textbf{CVR Increase}: CVR up; CVR rose; CVR increased; CVR improved; CVR higher.
\item \textbf{CVR Decrease}: CVR down; CVR fell; CVR decreased; CVR declined; CVR lower.
\item \textbf{CPA Increase}: CPA up; CPA rose; CPA increased; CPA higher; Cost up.
\item \textbf{CPA Decrease}: CPA down; CPA fell; CPA decreased; CPA lower; Cost down.
\end{itemize}
\noindent\textcolor{AIGrayLine}{\rule{\linewidth}{0.35pt}}
\PromptSection{Strategy (7 categories $\times$ 5 variants)}
\begin{itemize}
\item \textbf{Conservative}: Conservative approach suggested; Lower bid indicated; Cautious bidding; Bid low; Conservative bid.
\item \textbf{Moderate}: Moderate approach suggested; Balanced bid indicated; Moderate bidding; Bid moderate; Mid-range bid.
\item \textbf{Aggressive}: Aggressive approach possible; Higher bid indicated; Competitive bidding; Bid high; Aggressive bid.
\item \textbf{High pValue}: pValue \{pvalue\} - strong; Strong signal: \{pvalue\}; pValue high: \{pvalue\}; Good signal \{pvalue\}; Favorable pValue \{pvalue\}.
\item \textbf{Low pValue}: pValue \{pvalue\} - weak; Weak signal: \{pvalue\}; pValue low: \{pvalue\}; Poor signal \{pvalue\}; Unfavorable pValue \{pvalue\}.
\item \textbf{Budget Low}: Budget: \{budget\}; Remaining: \{budget\}; Budget \{budget\}; Available \{budget\}; Budget at \{budget\}.
\item \textbf{Budget High}: Budget: \{budget\}; Remaining: \{budget\}; Budget \{budget\}; Available \{budget\}; Budget at \{budget\}.
\end{itemize}
}
\end{AIBox}
\caption{Prompt templates for Concise style (high-conversion prompt-variant study): minimal tokens with essential information only.}
\label{fig:prompt_concise}
\end{figure*}

\begin{figure*}[th!]
\begin{AIBox}[title={Prompt Templates: Verbose Style}]
\parbox[t]{\textwidth}{
\small
\PromptSection{Task Description (4 variants)}
\begin{enumerate}
\item Your objective is to optimize the bidding strategy to achieve the target cost per acquisition of \{cpa\}, while balancing conversion volume and cost efficiency throughout the campaign.
\item The goal is to maintain cost per acquisition at the target level of \{cpa\}, ensuring that both conversion volume and cost efficiency are appropriately balanced based on current market conditions.
\item Optimize bidding decisions to meet the CPA target of \{cpa\}, taking into account the trade-off between maximizing conversions and maintaining cost efficiency.
\item Your task is to determine optimal bid levels that will achieve the target CPA of \{cpa\}, while considering both the opportunity to capture conversions and the need to control costs.
\end{enumerate}
\noindent\textcolor{AIGrayLine}{\rule{\linewidth}{0.35pt}}
\PromptSection{History (7 categories, representative examples)}
\begin{itemize}
\item \textbf{ROI Low}: The return on investment from the previous bidding action decreased, indicating that the relationship between costs incurred and conversions achieved was unfavorable.
\item \textbf{ROI Moderate}: The return on investment from the previous bidding action was moderate, indicating that the relationship between costs and conversions was within acceptable parameters.
\item \textbf{ROI Good}: The return on investment from the previous bidding action was high, indicating that the relationship between costs incurred and conversions achieved was very favorable.
\item \textbf{CVR Increase}: The conversion rate increased following the bidding action, indicating that the proportion of opportunities that resulted in conversions improved.
\item \textbf{CVR Decrease}: The conversion rate decreased following the bidding action, indicating that the proportion of opportunities that resulted in conversions declined.
\item \textbf{CPA Increase}: The cost per acquisition increased following the bidding action, indicating that the average cost to achieve each conversion rose.
\item \textbf{CPA Decrease}: The cost per acquisition decreased following the bidding action, indicating that the average cost to achieve each conversion fell.
\end{itemize}
\noindent\textcolor{AIGrayLine}{\rule{\linewidth}{0.35pt}}
\PromptSection{Strategy (7 categories, representative examples)}
\begin{itemize}
\item \textbf{Conservative}: Based on the current state indicators, including budget levels, signal quality, and recent performance metrics, a conservative bidding approach may be warranted to manage risk and preserve resources.
\item \textbf{Moderate}: Based on the current state indicators, including budget levels, signal quality, and recent performance metrics, a moderate bidding approach may be warranted to balance risk and opportunity.
\item \textbf{Aggressive}: Based on the current state indicators, including budget levels, signal quality, and recent performance metrics, there may be room for higher bidding to capture additional opportunities.
\item \textbf{High pValue}: The current pValue of \{pvalue\} indicates strong signal quality, suggesting that the predicted conversion probability for available opportunities is relatively high.
\item \textbf{Low pValue}: The current pValue of \{pvalue\} indicates weak signal quality, suggesting that the predicted conversion probability for available opportunities is relatively low.
\item \textbf{Budget Low/High}: Current budget remaining stands at \{budget\} of the original allocation, which represents the available resources for continued bidding activities.
\end{itemize}
}
\end{AIBox}
\caption{Prompt templates for Verbose style (high-conversion prompt-variant study): detailed explanations with full context and reasoning.}
\label{fig:prompt_verbose}
\end{figure*}

\begin{figure*}[th!]
\begin{AIBox}[title={Prompt Templates: Structured Style}]
\parbox[t]{\textwidth}{
\small
\PromptSection{Task Description (4 variants)}
\begin{enumerate}
\item Objective: Achieve CPA target of \{cpa\}.
\item Goal: Maintain CPA at \{cpa\}. Focus: Balance conversions and costs.
\item Target: CPA \{cpa\}. Task: Optimize bidding strategy.
\item CPA Target: \{cpa\}. Objective: Optimize bid decisions.
\end{enumerate}
\noindent\textcolor{AIGrayLine}{\rule{\linewidth}{0.35pt}}
\PromptSection{History (7 categories, representative examples)}
\begin{itemize}
\item \textbf{ROI Low}: Previous Performance: - ROI: Decreased - Efficiency: Below target
\item \textbf{ROI Moderate}: Previous Performance: - ROI: Moderate - Efficiency: Acceptable
\item \textbf{ROI Good}: Previous Performance: - ROI: High - Efficiency: Above target
\item \textbf{CVR Increase}: Conversion Trend: - CVR: Increased - Direction: Upward
\item \textbf{CVR Decrease}: Conversion Trend: - CVR: Decreased - Direction: Downward
\item \textbf{CPA Increase}: Cost Metrics: - CPA: Increased - Efficiency: Decreased
\item \textbf{CPA Decrease}: Cost Metrics: - CPA: Decreased - Efficiency: Increased
\end{itemize}
\noindent\textcolor{AIGrayLine}{\rule{\linewidth}{0.35pt}}
\PromptSection{Strategy (7 categories, representative examples)}
\begin{itemize}
\item \textbf{Conservative}: Recommendation: Conservative approach. Reason: Current signals suggest caution.
\item \textbf{Moderate}: Recommendation: Moderate approach. Reason: Current signals suggest balance.
\item \textbf{Aggressive}: Recommendation: Aggressive approach possible. Reason: Current signals suggest opportunity.
\item \textbf{High pValue}: Signal Quality: - pValue: \{pvalue\} - Strength: High - Assessment: Favorable
\item \textbf{Low pValue}: Signal Quality: - pValue: \{pvalue\} - Strength: Low - Assessment: Unfavorable
\item \textbf{Budget Low}: Budget Status: - Remaining: \{budget\} - Utilization: \{spent\} - Available: Limited
\item \textbf{Budget High}: Budget Status: - Remaining: \{budget\} - Utilization: \{spent\} - Available: Sufficient
\end{itemize}
}
\end{AIBox}
\caption{Prompt templates for Structured style (high-conversion prompt-variant study): label-value pairs with explicit field names.}
\label{fig:prompt_structured}
\end{figure*}

\end{document}